\documentclass[lettersize,journal]{IEEEtran}
\usepackage{amsmath,amsfonts}
\usepackage{array}
\usepackage[caption=false,font=normalsize,labelfont=sf,textfont=sf]{subfig}
\usepackage{textcomp}
\usepackage{stfloats}
\usepackage{url}
\usepackage{verbatim}
\usepackage{graphicx}
\usepackage{cite}
\usepackage{url}
\usepackage{hyperref}
\usepackage{multirow}
\usepackage{xcolor}
\usepackage{booktabs} 
\usepackage{makecell}
\usepackage[ruled,vlined]{algorithm2e}
\usepackage[table]{xcolor}
\begin{document}

\title{PL-VIWO2: A Lightweight, Fast and Robust Visual-Inertial-Wheel Odometry Using Points and Lines}

\author{Zhixin Zhang, Liang Zhao, and Pawel Ladosz
\thanks{This work was partially funded by the Robotics and AI Collaboration (RAICo). (Corresponding author: Pawel Ladosz).}
\thanks{Zhixin Zhang is with the Department of Electrical and Electronic Engineering, University of Manchester, Manchester M13 9PL, U.K.
Email: {\tt\small zhixin.zhang@manchester.ac.uk}}
\thanks{Liang Zhao is with the School of Informatics, University of Edinburgh, Edinburgh EH8 9AB, U.K.
Email: {\tt\small liang.zhao@ed.ac.uk}}
\thanks{Pawel Ladosz is with the Department of Mechanical, Aerospace and Civil Engineering, University of Manchester, Manchester M13 9PL, U.K.
Email: {\tt\small pawel.ladosz@manchester.ac.uk}}
}


\maketitle

\begin{abstract}
Vision-based odometry has been widely adopted in autonomous driving owing to its low cost and lightweight setup; however, its performance often degrades in complex outdoor urban environments. To address these challenges, we propose PL-VIWO2, a filter-based visual–inertial–wheel odometry system that integrates an IMU, wheel encoder, and camera (supporting both monocular and stereo) for long-term robust state estimation. The main contributions are: (i) a novel line feature processing framework that exploits the geometric relationship between 2D feature points and lines, enabling fast and robust line tracking and triangulation while ensuring real-time performance; (ii) an SE(2)-constrained SE(3) wheel pre-integration method that leverages the planar motion characteristics of ground vehicles for accurate wheel updates; and (iii) an efficient motion consistency check (MCC) that filters out dynamic features by jointly using IMU and wheel measurements. Extensive experiments on Monte Carlo simulations and public autonomous driving datasets demonstrate that PL-VIWO2 outperforms state-of-the-art methods in terms of accuracy, efficiency, and robustness.
\end{abstract}


\section{Introduction}
\IEEEPARstart{T}{he} complex nature of urban environments makes it difficult to ensure long-term stable state estimation in autonomous driving, which continues to attract considerable research attentions. Visual-inertial Navigation Systems (VINS) or Visual-inertial-Odometry (VIO), which integrate cameras and inertial measurement units (IMUs), have been widely regarded as a promising solution for state estimation in unmanned systems due to their low cost and complementary sensing characteristics. Over the past decade, numerous representative VIO have been proposed, including ORB-SLAM3~\cite{c1}, VINS-Mono~\cite{c2}, and OpenVINS~\cite{c3}. 

However, the IMU often fails to provide accurate measurements under high-speed motion, such as highway driving, leading to inaccurate estimation~\cite{c4}. Furthermore, VINS may exhibit additional unobservable degrees of freedom during specific motion patterns, further degrading its performance in autonomous driving scenarios~\cite{c5}. These challenges can be effectively addressed by incorporating wheel encoder, a standard sensor on ground vehicles, which provides additional observation information~\cite{c6,c7}. Since the wheel measurements only provide 2D motion information, while the visual-inertial-wheel odometry (VIWO) system estimates a full 3D state, the planar motion characteristics of ground vehicles can be exploited to bridge this gap. Specifically, the 2D wheel measurements are extended into an SE(2) constricted SE(3) state space under planar motion assumption, thereby improving estimation accuracy.
\begin{figure}
    \centering
    \includegraphics[width=0.95\linewidth]{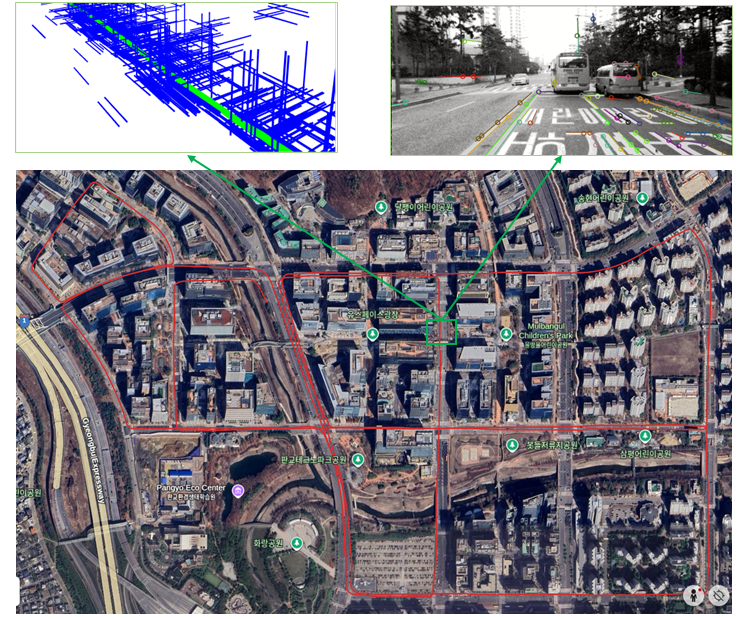}
    \caption{PL-VIWO2 (Stereo setup) test in KAIST Urban38.\textbf{Bottom}: Estimated trajectory (red line) of PL-VIWO2 on KAIST Urban38 (total length: 11.4 km), visualized in Google Earth. 
    \textbf{Top-Left}: Line triangulation results, where blue lines represent the spatial line features while green line is the trajectory. 
    \textbf{Top-Right}: Point–line pairing results, where circles with the same color as a line indicate point features lying on the corresponding line feature.}
    \label{fig:first-fig}
\end{figure}

For vision-based odometry, point features are commonly tracked across different camera poses and constrain the system state by using descriptors~\cite{c8} or optical flow~\cite{c9} methods. However, acquiring sufficiently reliable features in urban environments remains challenging due to dynamic objects, poor lighting conditions, and texture-less surfaces. To address this, incorporating additional geometric features, such as lines, has been proven to be effective in previous works~\cite{c10, c11, c12}. Despite its potential, the application of line features in autonomous driving scenarios faces several limitations that hinder their practical deployment. First, processing line features introduces additional computational overhead, which may compromise the real-time performance. Secondly, conventional line tracking methods typically rely on descriptors, which often perform poorly in complex outdoor environments. For instance, high-speed linear motion or false line detections can lead to unreliable feature correspondences between views. Thirdly, due to the dominant planar motion of ground vehicles, classical triangulation methods may suffer from degenerate motion, resulting in inaccurate triangulation results and ultimately degrading the state estimation accuracy~\cite{c13}.

Another challenge for state estimation in autonomous driving environments is dynamic objects, such as moving vehicles and pedestrians. The inclusion of dynamic feature points can significantly degrade estimation accuracy. To mitigate this issue, some approaches utilize deep learning techniques to perform semantic segmentation and filter out dynamic points~\cite{c14,c15}. But such methods are often computationally demanding and may not guarantee real-time performance on resource-limited platforms. Therefore, this paper leverages a lightweight and efficient Motion Consistency Check (MCC) by using the measurement information of IMU and wheel to reject dynamic features. Specifically, the motion of each feature point within a sliding window is compared against the predicted motion derived from inertial and wheel data. Features exhibiting inconsistent motion patterns are classified as dynamic or outliers and further excluded from state estimation. 

To address the aforementioned challenges, we propose a novel visual-inertial-wheel odometry (VIWO) system that integrates both point and line features, named PL-VIWO2, designed for long-term robust and accurate state estimation of autonomous driving in complex outdoor environments. The main contributions of this work are:
\begin{itemize}
\item A lightweight, fast, and robust VIWO system leveraging both point and line features is proposed for handling complex autonomous driving scenarios, with support for both monocular and stereo configurations.
\item Two novel optical-flow-based line feature tracking methods are developed and combined to enable efficient and reliable line tracking in complex outdoor environments.
\item A robust and accurate 3D line triangulation pipeline exploiting the geometric relationship between point and line features is implemented, including both initialization and optimization-based refinement.
\item An $\mathrm{SE}(2)$-constrained $\mathrm{SE}(3)$ wheel pre-integration method is designed based on the planar motion characteristics of vehicles for wheel updates.
\item A motion consistency check (MCC) module that eliminates the influence of dynamic features by leveraging IMU and wheel measurements.
\item Validation through both Monte Carlo simulations and real-world experiments, demonstrating robustness, efficiency, and accuracy.
\item The source code of the system will be released as open source upon acceptance of the paper.
\end{itemize}

Compared with our previous publication PL-VIWO~\cite{c16}, the new contributions of PL-VIWO2 are:
\begin{itemize}
    \item PL-VIWO2 is extended to support both monocular and stereo configurations, ensuring compatibility with various hardware setups, whereas PL-VIWO supports only the monocular configuration.
    \item A second-stage optical-flow-based line matching method for tracking line features that cannot be directly tracked using point tracking results.
    \item An optimization refinement strategy is introduced after 3D line initialization to further improve the accuracy of line triangulation results.
    \item Extensive evaluations are conducted on both simulations and real-world datasets, demonstrating that PL-VIWO2 achieves superior robustness and accuracy compared with PL-VIWO and other SOTA methods.
\end{itemize}

The rest of this paper is organized as follows. Section~\ref{sec:related_work} reviews related work. Section~\ref{sec:system_overview} introduces the system overview and the state estimation method that integrates measurements from three sensors. Section~\ref{sec:visual_frontend} details the proposed novel visual processing pipeline, including the point-based line feature tracking, triangulation, stereo configuration and motion consistency check. Section~\ref{sec:simulation} presents Monte Carlo simulation results and evaluates the proposed line triangulation methods. Section~\ref{sec:experiments} reports real-world experimental results and performance evaluation. Finally, Section~\ref{sec:conclusion} concludes the paper.

\section{Related Works}
\subsection{Visual-inertial-Wheel Odometry}
Ground vehicles serve as a common platform for autonomous mobile robots, with localization being a long-standing research focus. The fusion of visual-inertial navigation systems (VINS) and wheel encoder data has emerged as a widely adopted approach to enhance the accuracy and robustness of state estimation. It has been shown that VINS may suffer from additional unobservable degrees of freedom under certain motion patterns, and incorporating wheel encoder measurements can help maintain system consistency~\cite{c5}.
According to the fusion strategy, VIWO systems can generally be categorized into two types: filter-based methods~\cite{c5,c7,c17} and graph optimization-based methods~\cite{c18,c19,c20}. Filter-based approaches are typically faster and more computationally efficient, while optimization-based methods offer higher accuracy at the cost of increased computational complexity.

Moreover, the motion of ground vehicles typically adheres to planar constraints, and this prior knowledge can be leveraged to improve the accuracy of state estimation. For instance, ~\cite{c21,c22} proposed odometry systems that parameterize the robot pose within an SE(2)-constrained SE(3) framework. \cite{c23} improved estimation accuracy by leveraging camera-to-ground geometric parameters as constraints. Due to the high noise often present in angular velocity measurements from wheel encoders, \cite{c6,c19} utilize the angular velocity pre-integration from IMU measurements and the linear velocity from wheel encoders as inputs to the system. In this paper, we propose a SE(2)-constrained SE(3) pre-integration using wheel speed measurements, and incorporate the result into a Kalman filter for state update. This approach not only reinforces the constraints on degrees of freedom outside the SE(2) manifold but also maximizes the utilization of information from both the IMU and wheel encoders.

\subsection{Line-aided VIO}
Incorporating line features into VIO or VI-SLAM systems effectively improves localization accuracy by adding geometric constraints~\cite{c10,c11,c12}. Like point features, lines must be tracked across frames. Classical methods typically use the Line Segment Detector (LSD)~\cite{c25} and descriptors like LBD~\cite{c26}, but descriptor-based tracking is often computationally expensive, thus unsuitable for real-time use. To address this,\cite{c27} proposed a faster variant of LSD for real-time detection. Based on their method, \cite{c28} added point on Line constraints and parallel line constraints to further improve the estimation accuracy. Optical flow-based line tracking methods have also been explored to improve efficiency~\cite{c29,c30}. In AirVO~\cite{c31}, a point-assisted strategy was proposed, using point correspondences to aid line tracking. AirSLAM~\cite{c32} extended this with a learning-based model that jointly detects and tracks points and lines, still relying on point matches. However, it targets indoor scenes and struggles outdoors. Inspired by these ideas, we design a lightweight approach that leverages point tracks for fast line tracking and uses optical flow to refine unmatched lines, improving robustness in outdoor environments.

Another key challenge lies in line triangulation. Due to degenerate motions, e.g. straight driving with line direction, triangulating lines (like lane markings or roadside edges) often leads to poor accuracy~\cite{c13}. This issue is prominent for ground vehicles. \cite{c33} addressed this by applying parallel constraints to handle degenerate lines. Other approaches avoid triangulation altogether, they use line directions based on Manhattan assumptions as structural priors~\cite{c34,c35,c36}. In this work, we propose a new triangulation pipeline that leverages 2D geometric relationships between points and lines to robustly recover 3D line structures, enabling efficient and reliable triangulation initialization even under degenerate conditions. After that, a 4DOF optimization refinement is performed based on all the information.

\subsection{VIO in Dynamic Environments}
The deployment of unmanned systems in real-world scenarios inevitably involves dynamic environments, where achieving fast and accurate localization remains a core challenge~\cite{c37}. Deep learning-based semantic segmentation proved to be effective for removing dynamic objects~\cite{c14,c15}; however, such methods are computationally expensive and often unsuitable for real-time applications on resource-limited platforms. As an alternative, motion consistency check (MCC) offers a lightweight and efficient solution. For example,\cite{c38,c39} reject feature points with large deviations from IMU integration, while\cite{c40} uses RGB-D data to label 3D points as dynamic if they violate motion consistency. In wheel-integrated systems, projected errors with respect to wheel odometry can also serve as a criterion for dynamic point rejection~\cite{c19}. However, these approaches typically rely on a single sensor. In contrast, PL-VIWO2 employs a fused motion reference derived from both IMU and wheel speed, enabling more reliable detection of dynamic features. Moreover, MCC is applied to both point and line features, further enhancing system robustness and accuracy in dynamic environments.

\section{System Overview and State Estimation}
\label{sec:system_overview}
In this section, we first present the overall system framework. We then provide a detailed description of the sensor fusion methodology, including IMU propagation, camera-feature updates, and wheel encoder updates.
\label{sec:related_work}
\begin{figure}[t]
    \centering
    \includegraphics[width= 0.5\textwidth]{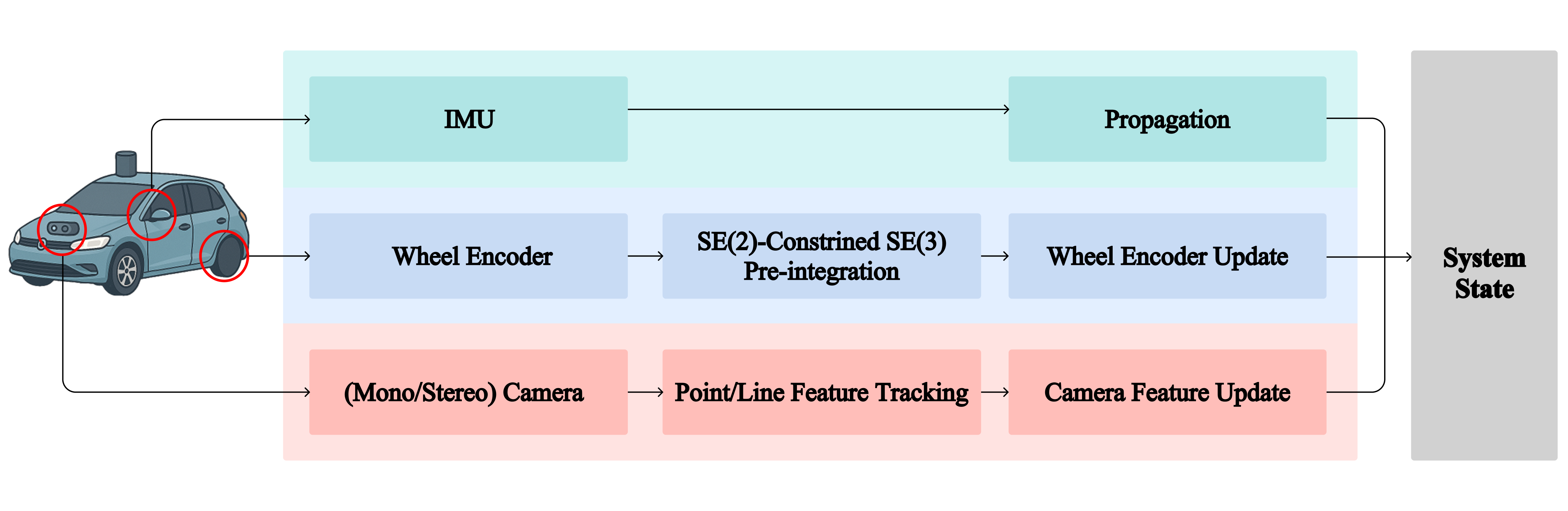}
    \caption{The overall framework of PL-VIWO2 system is divided into three components, each represented by different color-coded boxes.} 
    \label{Framework}
\end{figure}
\subsection{System Overview}
The system state vector $\textbf{x}_{k}$ at time step $k$ consists of current IMU state $\textbf{x}_{I_{k}}$ and  $n$ historical IMU clones $\textbf{x}_{H_{k}}$. 
\begin{align}
\textbf{x}_{k} &= (\textbf{x}_{I_{k+1}}, \textbf{x}_{H_{k}}) \\
\textbf{x}_{I_{k}} &= (^{I_{k}}_{G}\textbf{R}, \ ^{G}\textbf{p}_{I_k}, \ ^{G}\textbf{v}_{I_k}, \ \textbf{b}_g, \ \textbf{b}_a) \\
\textbf{x}_{H_{k}} &= (^{I_{k-1}}_{G}\textbf{R}, \ ^{G}\textbf{p}_{I_{k-1}}, \ \dots, \ ^{I_{k-n}}_{G}\textbf{R}, \ ^{G}\textbf{p}_{I_{k-n}})
\end{align}
where $^{B}_{A}\textbf{R}$ is the rotation matrix from frame $A$ to $B$ and $ ^{A}\textbf{p}_{B}$ is position of $B$ in $A$. The terms $\textbf{b}_g$ and $\textbf{b}_a$ denote the gyroscope and accelerometer bias, respectively. In this work, we define the frames as follows: $G$ denotes the global frame, $I$ the IMU frame, $C$ the camera frame and $W$ the wheel odometry frame.

The overall framework of proposed PL-VIWO2 is illustrated in Fig.~\ref{Framework}. The system consists of three main components: IMU, Wheel Encoder, and Camera. Inherited from the framework of MINS \cite{c4}, the system uses the IMU measurements for state propagation, while wheel encoder and camera measurements are incorporated as observations to update system state.

\subsection{IMU Propagation}
The IMU measures the local acceleration ${}^I\textbf{a}_{m_{k}}$ and angular velocity ${}^I\textbf{w}_{m_{k}}$ of the system at a high frequency, which is used for propagation in the proposed VIWO system.
\begin{align}
    {}^I\boldsymbol{w}_{m_{k}} &= {}^I\boldsymbol{w}_{k} + \mathbf{b}_{g_{k}} + \mathbf{n}_{g_{k}} \\
    {}^I\mathbf{a}_{m_{k}} &= {}^I\mathbf{a}_{k} + {}^I_G\mathbf{R}_{k}\,{}^G\mathbf{g} + \mathbf{b}_{a_{k}} + \mathbf{n}_{a_{k}}
\end{align}
where ${}^I\boldsymbol{w}_{m_{k}}$ and ${}^I\mathbf{a}_{m_{k}}$ denote the true linear acceleration and angular velocity, respectively. $\mathbf{n}_{g_{k}}$ and $ \mathbf{n}_{a_{k}}$ are the measurement noise of the gyroscope and accelerometer, respectively. The system state can be propagated from $t_k$ to $t_{k+1}$ using IMU measurements. 
\begin{equation}
\hat{\textbf{x}}_{I_{k+1}} = \textbf{f}(\hat{\textbf{x}}_{I_{k}}, \  {}^I\textbf{w}_{m_{k}}, \  {}^I\textbf{a}_{m_{k}})
\label{propagation}
\end{equation}
By linearizing this propagation model, the IMU state covariance matrix ${\boldsymbol{\Sigma}}$ can be propagated as follows:
\begin{equation}
    \boldsymbol{\Sigma}_{k+1} = \boldsymbol{F}_{k} \boldsymbol{\Sigma}_{k} \boldsymbol{F}_{k}^{T} + \boldsymbol{G}_{k} \boldsymbol{Q} \boldsymbol{G}_{k}^{T}
\end{equation}
where $\boldsymbol{F}_{k}$ and $\boldsymbol{G}_{k}$ are the Jacobians of $\mathbf{f}$ with respect to (w.r.t.) the IMU state and IMU noise, respectively. A detailed derivation can be found in \cite{c41}.

\subsection{Camera-Feature Update}
By tracking the image features across different camera poses, the feature observations and camera poses can be modeled as follows:
\begin{equation}
    \boldsymbol{z} = \boldsymbol{h}(\hat{\boldsymbol{x}}_{I_{k}}, {}^{G}\hat{\boldsymbol{x}}_{f}) + \boldsymbol{n}_{z}
\end{equation}
where $\boldsymbol{n}_{z}$ is the camera observation noise and ${}^{G}\boldsymbol{x}_{f}$ represents the estimated point or line feature in the global frame. These two measurement models are described in the following.
\begin{figure}[t]
    \centering
    \subfloat[]{\includegraphics[width=0.23\textwidth]{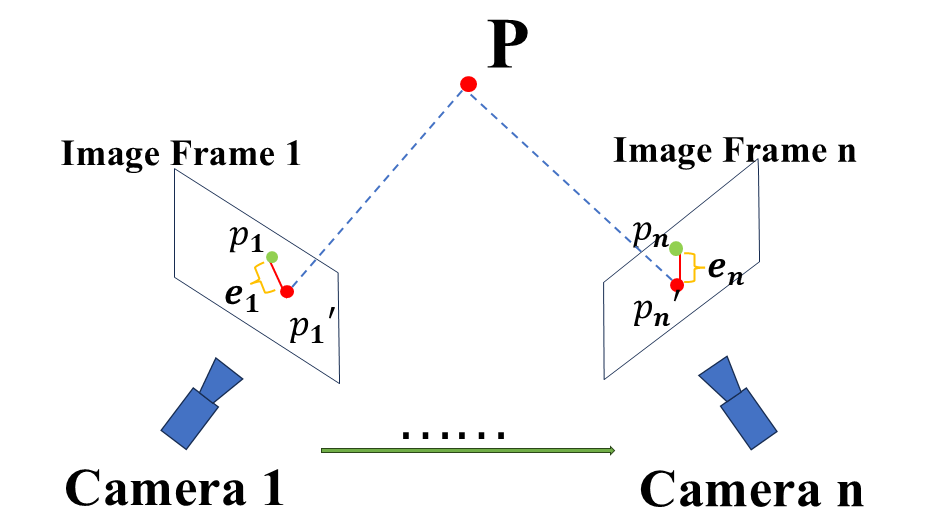}}
    \subfloat[]{\includegraphics[width=0.23\textwidth]{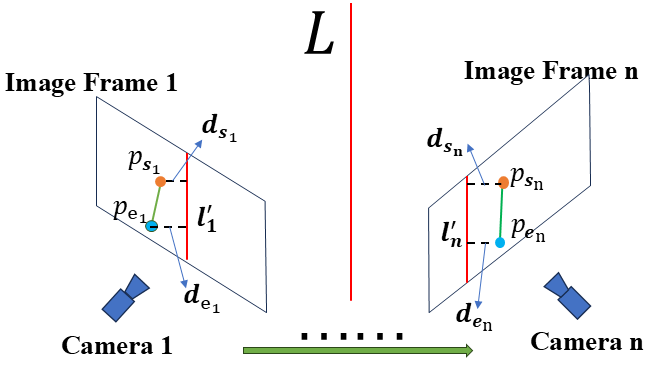}}
    \caption{The feature measurement model. (a) Point model; (b) Line model.}
    \label{fig:PiontMeasurementModel}
\end{figure}
\subsubsection{Point Feature Measurement Model}

The point feature measurement model describes how a 3D feature point is projected onto the image plane:
\begin{equation}
    \hat{\mathbf{z}} = \pi \big( {}^{C}\hat{\mathbf{p}}_f \big) + \mathbf{n}_z,
    \quad
    {}^{C}\mathbf{p}_f = {}^{C}_{I}\mathbf{R} \Big( {}^{I}_{G}\mathbf{R} ( {}^{G}\hat{\mathbf{p}}_f - {}^{G}\mathbf{p}_I ) + {}^{I}\mathbf{p}_C \Big)
\label{eq:campoint}
\end{equation}
where $\pi(\cdot)$ denotes the standard camera projection function, ${}^{C}\hat{\mathbf{p}}_f$ is the estimated feature position in the global frame, ${}^{I}_{C}\mathbf{R}$ and ${}^{C}\mathbf{p}_I$ are the known extrinsic rotation and translation from the IMU frame to the camera frame. Then, the error state observation is defined as the 2D reprojection error between the observed image feature point $\bar{\boldsymbol{z}}$ in Eq.~(\ref{eq:campoint}) and the projected point $\hat{\boldsymbol{z}}$, as shown in~Fig~\ref{fig:PiontMeasurementModel}(a):
\begin{equation}
    \tilde{\boldsymbol{z}} = \bar{\boldsymbol{z}} - \hat{\boldsymbol{z}}
\end{equation}

\subsubsection{Line Feature Measurement Model}
A spatial line represented in Plücker coordinates consists of two 3D vectors: the normal vector $\boldsymbol{n}$ and the direction $\boldsymbol{v}$:
\begin{equation}
    \textbf{L} = \left[\begin{array}{c c}
    \textbf{n} & \textbf{v}
\end{array}\right] ^\top
\label{eq:plucker}
\end{equation}

\begin{figure*}[t]
    \centering
    \includegraphics[width= 0.9 \textwidth]{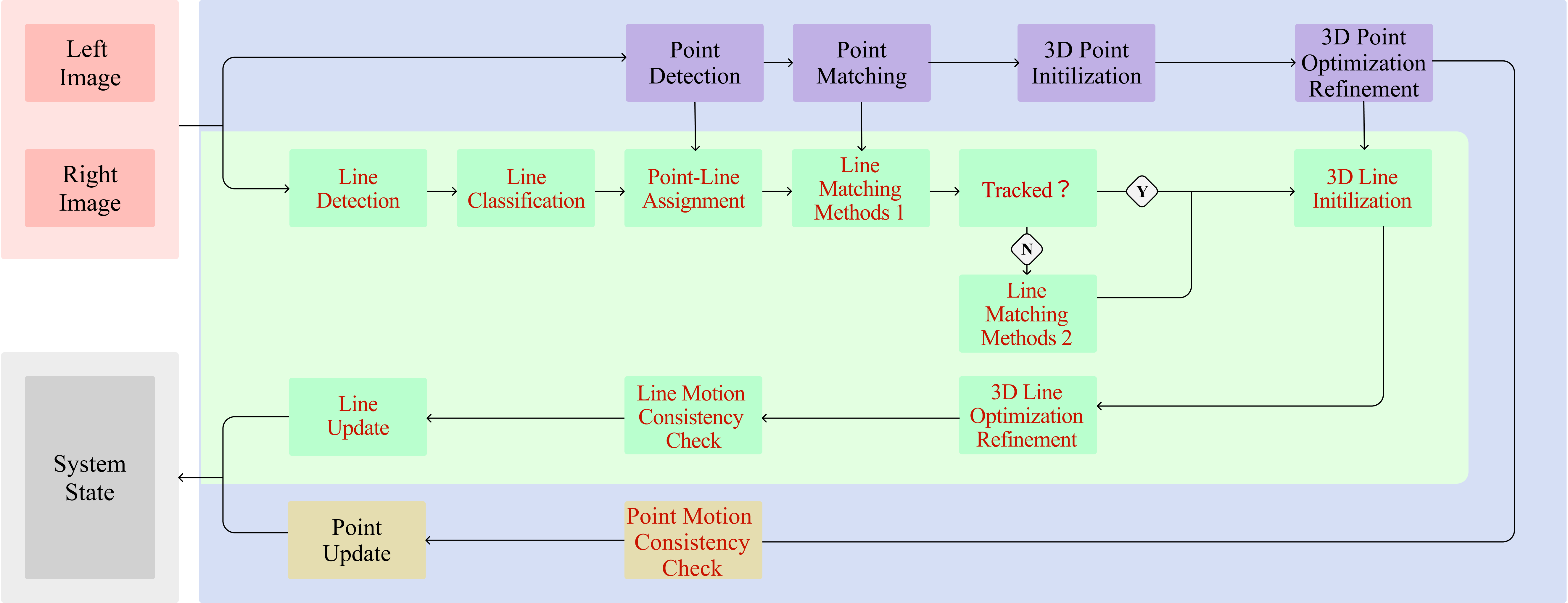}
    \caption{Flowchart of the proposed visual processing pipeline. The boxes highlighted with red text indicate the key contributions of this work.} 
    \label{Fig:Framework}
\end{figure*}
Similar to the point feature measurement model, a 3D line feature represented by Plücker coordinates is projected onto the image plane using the line projection function:
\begin{equation}
\begin{aligned}
\mathbf{l} &= \left[\begin{array}{c c}
\mathbf{K}_{L} & \mathbf{0}_3
\end{array}\right]{^{C}\mathbf{L}}
= \left[\begin{array}{c c}
\mathbf{K}_{L} & \mathbf{0}_3
\end{array}\right] \left[\begin{array}{c c} ^{C}{\mathbf{n}} & ^{C}{\mathbf{v}} \end{array}\right]^{\top}\\
&= \left[\begin{array}{c c} \mathbf{K}_{L} & \mathbf{0}_3 \end{array}\right]
\left[\begin{array}{c c} _{G}^{C}{\mathbf{R}} & [^{C}{\mathbf{p}}_{G}]_\times {_{G}^{C}{\mathbf{R}}}\\ \mathbf{0}_3 & _{G}^{C}{\mathbf{R}} \end{array}\right]
\left[\begin{array}{c c} ^{G}{\mathbf{n}} & ^{G}{\mathbf{v}} \end{array}\right]^{\top}\\
\end{aligned}
\end{equation}
where $\mathbf{K}_{L}$ is camera intrinsic matrix for line projection and $[\cdot]_{\times}$ represents the skew-symmetric matrix. The observation error is defined as the distance between the observed two endpoints $\textbf{p}_{s}$ and $\textbf{p}_{e}$, to the projected line $\mathbf{l}$, as shown in~Fig.~\ref{fig:PiontMeasurementModel}(b):
\begin{equation}
\mathbf{z_l} = \left[\begin{array}{cc}
    \mathbf{d}_{s} & \mathbf{d}_{e} 
\end{array}\right]^{\top} =
\left[\begin{array}{c c}
\frac{\mathbf{p}_s^\top \mathbf{l}}{\sqrt{l_1^2 + l_2^2}} & \frac{\mathbf{p}_e^\top \mathbf{l}}{\sqrt{l_1^2 + l_2^2}}
\end{array}\right] ^\top
\label{eq:line_measurement}
\end{equation}
where $\mathbf{p_s} = \begin{bmatrix} u_s& v_s&1 \end{bmatrix}^\top$ and $\mathbf{p_e} = \begin{bmatrix} u_e& v_e&1 \end{bmatrix}^\top$ denote the homogeneous coordinates of start and end points of the 2D line segment.

\subsubsection{MSCKF Update}
In order to solve the dimensionality explosion problem caused by keeping visual features in the state vector, Multi-State-Constrains-Kalman-Filter (MSCKF) uses the null space to marginalize the features and only updates the system state \cite{c42}. For a camera feature measurement, the error state model can be generically represented as :
\begin{equation}
\tilde{\textbf{z}} = \bar{\textbf{z}} - \textbf{h}(\hat{\textbf{x}}_{I_{k}}, {}^{G}\hat{\textbf{x}}_{f})
\end{equation}
where $\tilde{\textbf{z}}$ is the error between observation and projected featureand $^{G}\hat{\textbf{x}}_{f}$ is the estimated point or line feature in the global frame. Then this error modal can be linearized as:
\begin{equation}
    \tilde{\textbf{z}} = \textbf{H}_{x} \tilde{\textbf{x}}_{I_{k}} + \textbf{H}_{f} {}^{G}\tilde{\textbf{x}}_{f} + \textbf{n}_{z}
\end{equation}
where $\textbf{H}_x$ and $\textbf{H}_f$ are the Jacobians w.r.t. system state $\textbf{x}_{I_{k}}$ and feature $^{G}\textbf{x}_{f}$, respectively. And $\textbf{n}_{z}$ is the measurement noise.  By decomposing $\textbf{H}_f$, a new measurement modal independent of feature error can be derived :
\begin{equation}
\begin{aligned}
    \tilde{\textbf{z}}' &= \textbf{Q}_{n}^{T} \tilde{\textbf{z}} 
    \\ &= \textbf{Q}_{n}^{T} \textbf{H}_x \tilde{\textbf{x}} + \textbf{Q}_{n}^{T} \textbf{H}_{f} {}^{G}\tilde{\textbf{x}}_{f} + \textbf{Q}_{n}^{T} \textbf{n}_{f}
    \\ &= \textbf{H}'_x \tilde{\textbf{x}}_{I_{k}} + \textbf{n}'_{z}
\end{aligned}
\end{equation}
where $\textbf{Q}_{n}^{T}$ is the null space of $\textbf{H}_{f}$ and $\textbf{Q}_{n}^{T} \textbf{H}_{f} = \textbf{0}$. Then the standard EKF update can be performed. 

\subsection{Wheel Encoder Update}
The wheel encoder measures the rotation rate of each wheel:
\begin{equation}
    {w}_{ml_{k}} = {w}_{l_k} + n_{w_{l}},
    \quad {w}_{mr_{k}} = {w}_{r_k} + n_{w_{r}}
\end{equation}
where ${w}_{m_{lk}}$ and ${w}_{m_{rk}}$ denote the angular velocity of the left and right wheel at time step $k$, respectively, and $n_{w_{l}}$ and $n_{w_{r}}$ represent the measurement noise. By using the radius of each wheel $r$ and baselink length $b$, the 2D linear velocity and angular velocity can be computed as follows:
\begin{equation}
{}^{{W}_{k}}w = \frac{(\omega_{r_{k}} r_r - \omega_{l_{k}} r_l)}{b}, 
\quad
{}^{W_{k}}v = \frac{(\omega_{r_{k}} r_r + \omega_{l_{k}} r_l)}{2}
\end{equation}
The wheel odometry only provides 2D pose information, and does not provide information to the remaining Z-axis, pitch and row. However, since autonomous driving often moves on a 2D plane, based on this plane assumption, it can be assumed that these three degrees of freedom are zero:
\begin{equation}
    \boldsymbol{v} =  \begin{bmatrix} {}^{W_{k}}v & 0 & 0 \end{bmatrix},
    \quad  \boldsymbol{w} =  \begin{bmatrix}0 & 0 & {}^{W_{k}}w \end{bmatrix}
\end{equation}
Then, an SE (2)-constrained SE (3) integration is implemented as wheel encoder observation, which describes the translation and rotation changes from time $t_{k-1}$ to $t_{k}$ in the world frame:
\begin{equation}
    \bar{\boldsymbol{z}}_{W} =  \left[\begin{array}{c c}
    \Delta ^{G}_{W}\boldsymbol{R} \\
    \Delta ^{G}\boldsymbol{p}_{W}
    \end{array}\right] =  \left[\begin{array}{c c}
    \\ 0
    \\ 0
    \\ \int_{t_{k-1}}^{t_k}{ {}^{W_{k}}w \ dt}
    \\ \int_{t_{k-1}}^{t_k}{{}^{W_{k}}v \ cos(\theta) \ dt}
    \\ \int_{t_{k-1}}^{t_k}{}{{}^{W_{k}}v \ sin(\theta) \ dt}
    \\ 0
    \end{array}\right]
\end{equation}
where $\theta$ denotes the local yaw angle, $ \Delta ^{G}_{W}\boldsymbol{R}$ and $\Delta ^{G}\boldsymbol{p}_{W}$ represent the 2D relative rotation and translation in the wheel odometry frame. This wheel measurement can also be estimated from IMU poses $({}^{I}_{G}\boldsymbol{R}, {}^{G}\boldsymbol{p}_{I})$ and the IMU-wheel extrinsic parameters  $({}^{I}_{W}\boldsymbol{R}, {}^{W}\boldsymbol{p}_{I})$:
\begin{equation}
\begin{aligned}
    \hat{\boldsymbol{z}}_{W} &= \boldsymbol{h}(\boldsymbol{x}_{k}) = 
    \left[\begin{array}{c c}
    Log({}^{W_{k}}_{G}\boldsymbol{R} \ {}^{G}_{W_{k-1}}\boldsymbol{R})
    \\ {}^{W_{k-1}}_{G}\boldsymbol{R}(^{G}\boldsymbol{p}_{W_{k}} - ^{G}\boldsymbol{p}_{W_{k-1}})
    \end{array}\right]
    \\ &=  \left[\begin{array}{c c}
    Log({}^{W}_{I}\boldsymbol{R} \ {}^{I_{k}}_{G}\boldsymbol{R} \ ({}^{W}_{I}\boldsymbol{R}\ {}_{G}^{I_{k-1}}\boldsymbol{R})^{T})
    \\ {}^{W}_{I}\boldsymbol{R}\ {}_{G}^{I_{k-1}}\boldsymbol{R} ((^{W}\boldsymbol{p}_{I_{k}} + {}^{W}_{I_{k}}{\boldsymbol{R}}) - (^{W}\boldsymbol{p}_{I_{k-1}} + {}^{W}_{I_{k-1}}{\boldsymbol{R}})
     \end{array}\right]
\end{aligned} 
\end{equation}
Similar to the camera update, the error between wheel observation $\boldsymbol{z}$ and its estimation $\hat{\boldsymbol{z}}$ is used for EKF update after linearization:
\begin{equation}
\begin{aligned}
   \tilde{\boldsymbol{z}}_w = \bar{\boldsymbol{z}} - \hat{\boldsymbol{z}} \approx \boldsymbol{H}_{W}\tilde{\boldsymbol{x}}_{k} + \boldsymbol{n}_{W}
\end{aligned}
\end{equation}
where $\boldsymbol{H}_{W}$ is the Jacobian w.r.t. the IMU state, and $\boldsymbol{n}_{W}$ is the observation noise. For the covariance matrix and Jacobian computation, we follow the procedure described in Algorithm 1 of~\cite{c7}. The modification is that we extend the observation noise to $6 \times 6$ and replace the original wheel measurement noise matrix $\mathbf{Q} \in \mathbb{R}^{6 \times 6}$ with:

\begin{equation}
\mathbf{Q} = 
\mathrm{diag}\!\left(
\sigma_{\omega}^2,\;
\sigma_{p}^2,\;
\sigma_{p}^2,\;
\sigma_{\omega}^2,\;
\sigma_{p}^2,\;
\sigma_{p}^2
\right)
\end{equation}
where $\sigma_{\omega}$ denotes the standard deviation of wheel encoder angular velocity measurement noise, and $\sigma_{p}$ characterizes the uncertainty of planar motion.

\section{Visual Process Pipeline}
\label{sec:visual_frontend}
This section presents the visual processing pipeline, as illustrated in Fig.~\ref{Fig:Framework}. Since the point update module is inherited from MINS~\cite{c4}, the focus here is on four proposed components: 2D line processing, line triangulation, stereo line implementation, and the motion consistency check.
\begin{figure*}[t]
    \centering
    \subfloat[]{\includegraphics[width=0.32\textwidth]{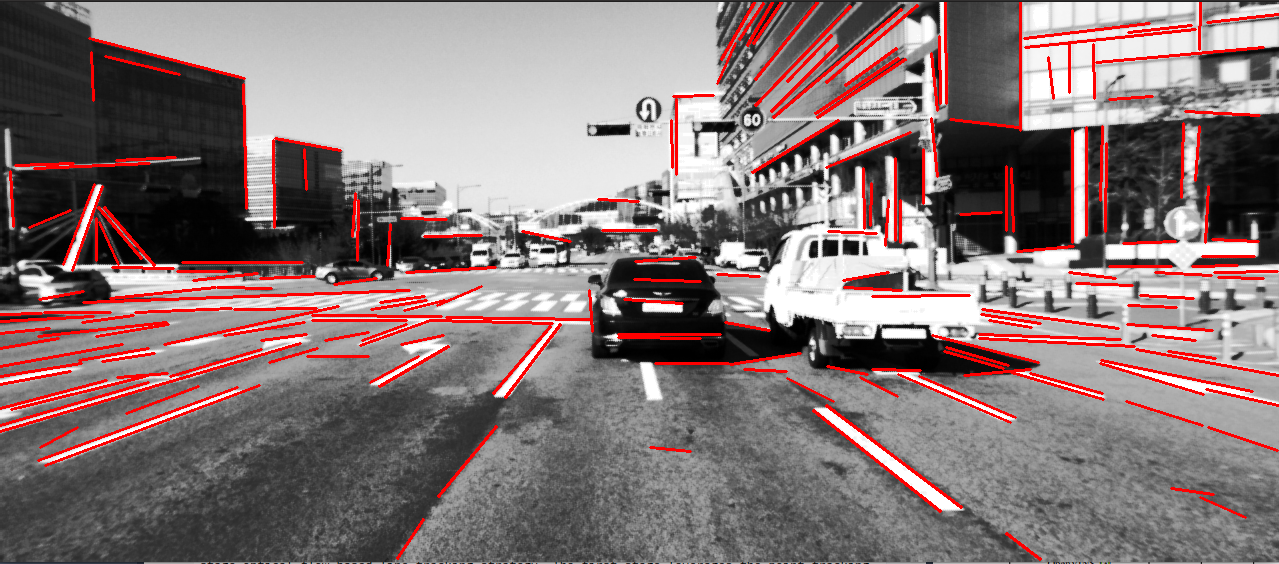}}
    \hspace{0.01\textwidth}
    \subfloat[]{\includegraphics[width=0.32\textwidth]{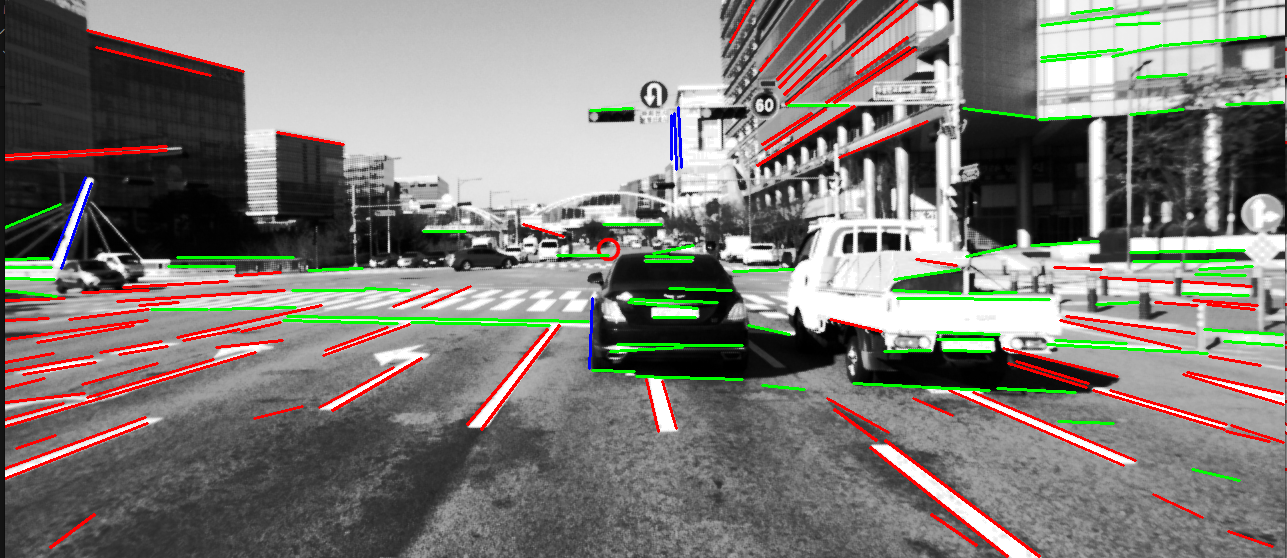}}
    \hspace{0.01\textwidth}
    \subfloat[]{\includegraphics[width=0.32\textwidth]{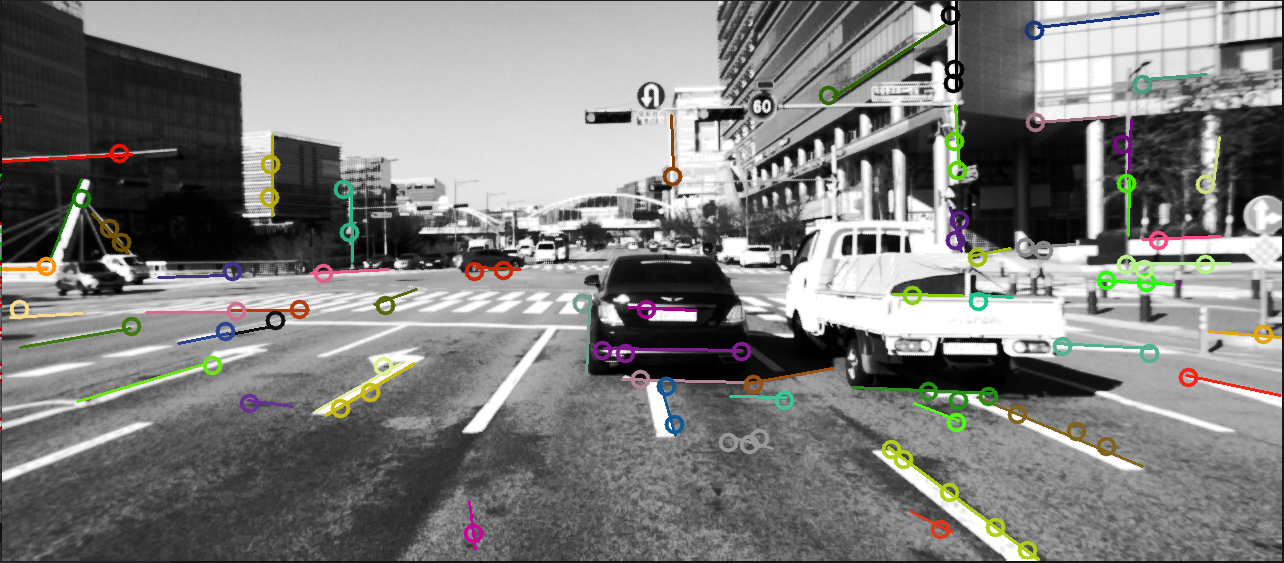}}
    \caption{2D Line process results in KAIST Urban 28. (a) Line detection result: red line segments represent the FLD results and lines which shorter than 30 pixels are filtered out for robustness; (b) Line classification result: lines are color-coded based on their alignment with the IMU frame axes — red for x-axis, green for y-axis, and blue for z-axis. Lines not aligned with any principal direction are excluded for clarity. The red circle near the image center indicates the vanishing point associated with the x-axis and vanishing points for y and z axes fall outside the image boundary; (c) Point line assignment result: circles lying on the same colored line represent point features assigned to that line.}
    \label{fig:2Dline}
\end{figure*}
\subsection{2D Line Feature Process}
\label{2DProcess}
When a sequence of camera images is fed into the system, the 2D processing of line features includes line detection, line classification, point-line assignment and line tracking.
\subsubsection{Line Detection}
The Fast Line Detector (FLD)\cite{c43} is adopted for 2D line segment detection due to its higher efficiency compared to the classical LSD\cite{c25}. To further accelerate the detection process, the original image is downsampled to one-quarter of its original resolution. However, FLD often splits a single physical line into multiple shorter segments, which negatively affects subsequent line matching accuracy. To address this issue, we adopt the line merging strategy from AirVO~\cite{c31}. An example of the FLD detection results is shown in Fig.~\ref{fig:2Dline}(a).

\subsubsection{Line Classification}
The detected straight lines often exhibit diverse orientations. However, since autonomous driving typically takes place in structured man-made environments, the scene generally conforms to the Manhattan world assumption, which facilitates more effective line classification~\cite{c44}. Specifically, line features are classified into four categories based on vanishing points: the x, y, and z directions relative to the IMU frame, along with an additional “none” category for lines that do not align with any of these principal axes.

A vanishing point is the image point at which parallel lines in 3D space appear to converge in the 2D projection. It is widely used to infer parallel line relationships in 3D from 2D images. The vanishing points along different axes can be computed as follows:
\begin{equation}
\begin{aligned}
    \textbf{vp}_{x} = \boldsymbol{\pi}({}_{I}^{C}\textbf{R} \begin{bmatrix} 1 & 0 & 0 \end{bmatrix}^\top) \\
    \textbf{vp}_{y} = \boldsymbol{\pi}({}_{I}^{C}\textbf{R} \begin{bmatrix} 0 & 1 & 0 \end{bmatrix}^\top) \\
    \textbf{vp}_{z} = \boldsymbol{\pi}({}_{I}^{C}\textbf{R} \begin{bmatrix} 0 & 0 & 1 \end{bmatrix}^\top)
\end{aligned}
\end{equation}
where ${}^{C}_{I}\mathbf{R}$ denotes the rotation matrix from the IMU frame to the camera frame, and $\boldsymbol{\pi}(\cdot)$ is the camera projection function. To classify a 2D line segment $\mathbf{l} = \begin{bmatrix}x_1 & y_1 & x_2 & y_2\end{bmatrix}$, we compute the similarity between the direction vector of the line and the vector from its midpoint to each vanishing point:
\begin{equation}
\text{s} = \left|
\frac{(x_2 - x_1,\ y_2 - y_1)}{\left| (x_2 - x_1,\ y_2 - y_1) \right|} \cdot
\frac{(\mathrm{vp}_x - x_m,\ \mathrm{vp}_y - y_m)}
{\left| (\mathrm{vp}_x - x_m,\ \mathrm{vp}_y - y_m) \right|}
\right|
\end{equation}
where $[x_m,\ y_m]$ is the midpoint of the line segment $\mathbf{l}$. After computing the similarity with each of the three vanishing points, the line is classified into the direction with the highest similarity score if the maximum score exceeds 0.97. The classification results are illustrated in Fig.~\ref{fig:2Dline}(b), and this directional information is utilized in the subsequent line triangulation process, helping to mitigate the impact of degenerate motions on line triangulation.

\subsubsection{Point-Line Assignment}
The relationship between detected point and line features in the 2D image can be established by computing the Euclidean distance from the point to the line segment. When the distance from a point feature to a line feature is below a predefined threshold, the point can be considered to lie on the corresponding line. Given a point feature $\mathbf{p} = {\begin{bmatrix} u_p & v_p \end{bmatrix}}^{\top}$ and a line feature $\mathbf{l} = \begin{bmatrix} u_s & v_s & u_e & v_e \end{bmatrix}^{\top}$ defined by its start and end points in the image plane, the point-to-line distance $d$ is computed as:
\[
\mathit{d} =
\begin{cases} 
\sqrt{(u_p - u_s)^2 + (v_p - v_s)^2}, & \text{if } \mathit{cross} \leq 0 \\[8pt]
\sqrt{(u_p - u_e)^2 + (v_p - u_e)^2}, & \text{if } \mathit{cross} > \mathit{len}^2 \\[8pt]
\frac{|(v_e - v_s)u_p + (u_s - u_e)v_p + (u_e v_s - u_s v_e)|}{\sqrt{(v_e - v_s)^2 + (u_s - u_e)^2}}, & \text{otherwise}
\end{cases}
\]
where $\mathit{cross}$ is the projection of the point on the line segment and $\mathit{len}$ is the length of line segment.
\begin{equation}
\begin{aligned}
    \mathit{len} &= \sqrt{{(u_s - u_e)^{2} + (v_s - v_e)^{2}}}  \\
    \mathit{cross} &= (u_e - u_s)(u_p - u_s) + (v_e - v_s)(v_p - v_s)
\end{aligned}
\end{equation}
If $\mathit{cross} < 0$, the projection of the point lies to the left of the start point $\mathbf{p}_s$. If $\mathit{cross} > \mathit{len}^2$, the projection lies beyond the end point $\mathbf{p}_e$. A 2D point is considered to be associated with a line segment only if its projection lies between the two endpoints and its perpendicular distance to the line is less than 3 pixels. This point-to-line association is used in subsequent line tracking and triangulation processes. The point line assignment result is shown in Fig.~\ref{fig:2Dline}(c). 

\subsubsection{Line Tracking}
\begin{figure*}
    \centering
    \subfloat[]{\includegraphics[width=0.9\textwidth]{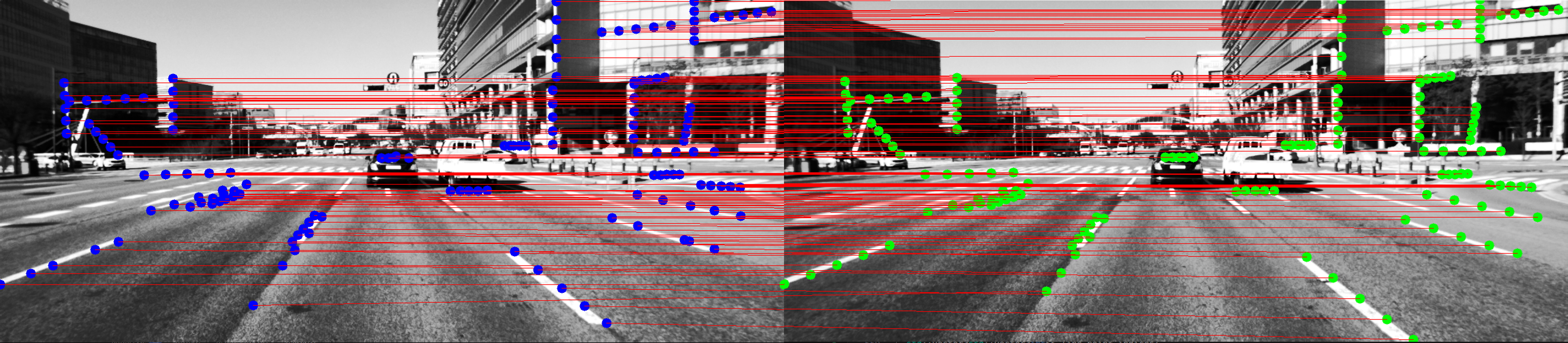}} \\
    \subfloat[]{\includegraphics[width=0.9\textwidth]{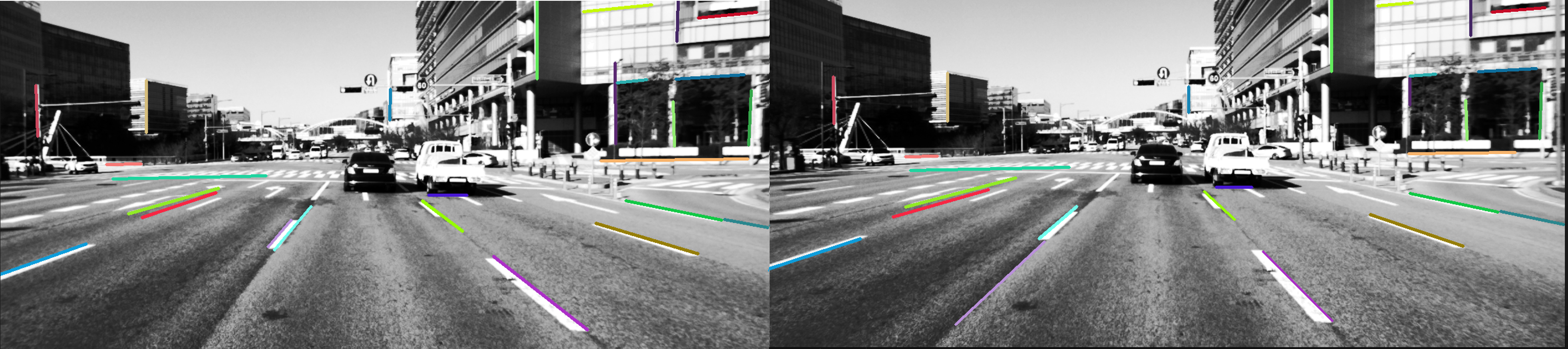}}
    \caption{(a) The second-stage line matching method. Five points are uniformly sampled along each line segment — two endpoints, two quarter points, and the midpoint. Blue circles (left) and green circles (right) denote the sampled points in two stereo frames. Points that fail to be tracked are omitted for clarity. Red lines represent the tracked correspondences between matched points.
    (b) Line matching results on a stereo image pair using the proposed two-stage method. Lines with the same color indicate successfully matched line segments between the two stereo frames.
    }
    \label{fig:matching}
\end{figure*}

To track line features across different images efficiently and accurately, we propose a two-stage optical flow-based line tracking strategy. The first stage leverages the point tracking results along with the point-to-line associations to establish initial line correspondences. Based on the number of associated point features, line segments are tracked according to the following rules:
\begin{itemize}
  \item Two line segments in adjacent frames are considered the same 3D line if they share at least two point features.
  \item If only one point is assigned, the differences in position and direction between the two line segments are calculated. If both differences are below the threshold, they are considered to be the same. (We assume a line maintains the same direction and undergoes limited positional changes between adjacent frames.)
\end{itemize}
After this initial stage, some line features may remain untracked due to missing or inaccurate point tracking. To address this, we introduce a second-stage line tracking method that directly tracks a set of sampled points along each untracked line. Specifically, five points are selected: the two endpoints, the midpoint, and two quarter points. These points are tracked across consecutive frames based on optical flow and then line tracking can be determined using the same rules as stage one. An example of this process is illustrated in Fig.~\ref{fig:matching}(a), while Fig.~\ref{fig:matching}(b) shows an example of matching results.

\subsection{3D Line Triangulation}
A 3D spatial line $\mathbf{L}$ can be expressed in two vectors, as shown in Eq.~(\ref{eq:plucker}). To obtain the 3D position of a line feature, an initial guess must first be generated, followed by an optimization step that using all the observation information to refine this estimation.
\begin{figure*}[t]
    \centering
    \subfloat[]{\includegraphics[width=0.3\textwidth]{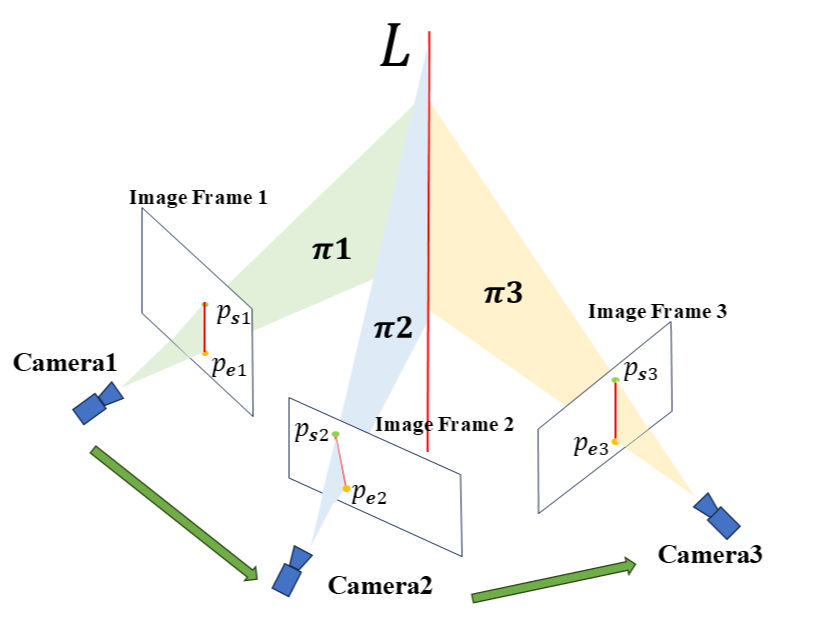}}
    \hspace{0.01\textwidth}
    \subfloat[]{\includegraphics[width=0.3\textwidth]{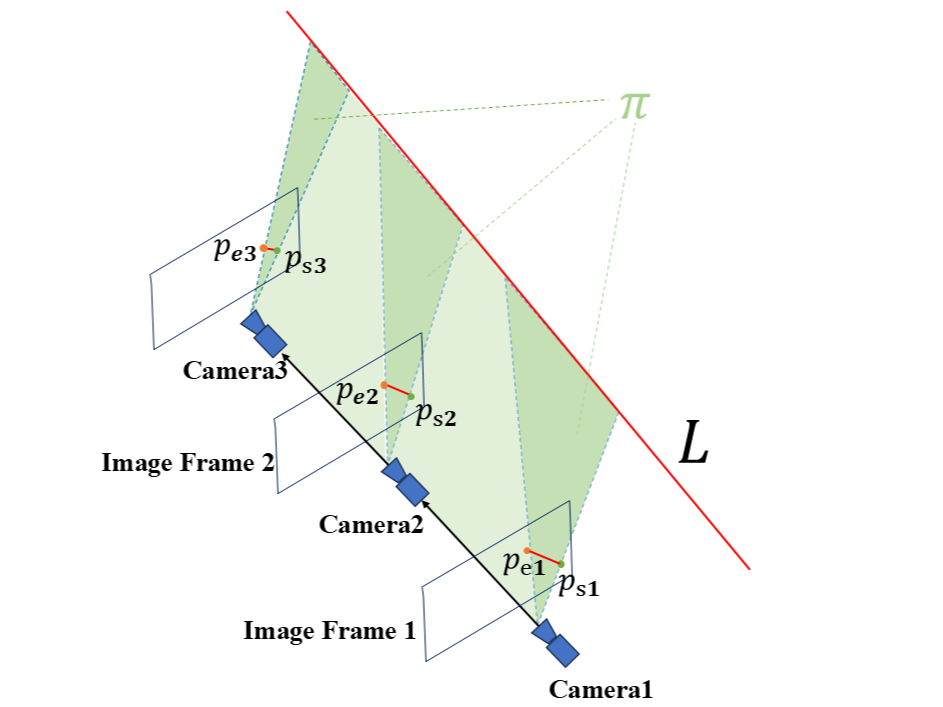}}
    \hspace{0.01\textwidth}
    \subfloat[]{\includegraphics[width=0.3\textwidth]{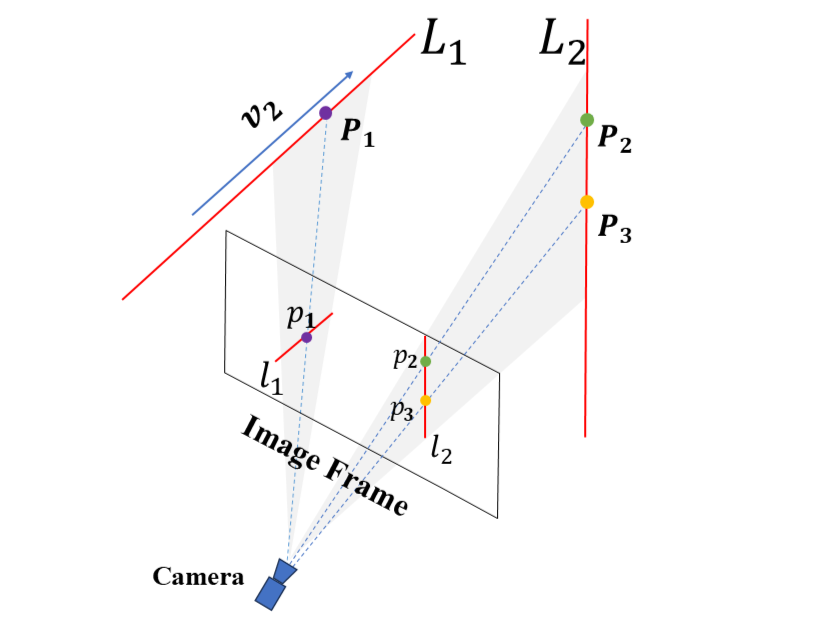}}
    \caption{ Line triangulation initialization
    (a) Triangulation from planes. 
    (b) Degenerate motion for line feature triangulation. Cameras stay in the same plane $\mathbf{\pi}$.
    (c) Triangulation from points and direction.}
    \label{fig:3D_line}
\end{figure*}
\begin{algorithm}[!t]
\caption{3D Line Initialization Strategy}
\label{alg:lineTriangulation}
\KwIn{2D Detected Lines, direction classification result, triangulated points on the line}
\KwOut{3D Line presented in Plücker coordinates}
\If{Direction Known}{
    \If{At least one triangulated point on the line}{
        Initialize with One Point and Known Direction\;
    }
}
\Else{
    \If{At least two triangulated points on the line}{
        Initialize with Two Points\;
    }
    \Else{
        Initialize with Plane Intersection\;
    }
}
\Return{Initial Guess of 3D Line}
\end{algorithm}
\subsubsection{3D Line Initialization}
Classical line triangulation initialization methods~\cite{c11,c12,c13} are typically based on plane intersection, where 2D line segments are back-projected into 3D space as planes, and the intersection of these planes defines the corresponding 3D line, as illustrated in Fig.~\ref{fig:3D_line}(a). However, such methods tend to fail when the camera moves along the line direction which is a scenario commonly encountered in vehicle navigation, like straight motion along lane or kerbstones, leading to poor initialization results, as shown in Fig.~\ref{fig:3D_line}(b). To overcome this issue, we proposed a robust line initialization pipeline that uses the point-on-line and line direction information introduced in Section~\ref{sec:visual_frontend}, which is summarized in Algorithm~\ref{alg:lineTriangulation}.

In general, three types of line initialization strategies are employed and Fig~\ref{fig:3D_line}(c) show the proposed first two methods. First, if a line has been successfully classified based on the 2D line classification and contains at least one triangulated feature point, its direction vector can be obtained as:
\begin{equation}
    \textbf{v} = {}_{G}^{I}\textbf{R} \textbf{u}
\end{equation}
where $\textbf{u}$ is the unit vector along the parallel axis. Then, the normal vector $\textbf{n}$ can be derived by using point $\textbf{p}$:
\begin{equation}
    \textbf{n} = \textbf{p} \times \textbf{v}
\end{equation}
If the direction of the line is unknown, but there are more than two triangulated points on the line, we choose to use the two 3D points $\textbf{p}_{1}$ and $\textbf{p}_{2}$ to initialize the line, as follows:
\begin{equation}
    \textbf{L} = \left[\begin{array}{c c}
    \textbf{n} \\
    \textbf{v}
    \end{array}\right] = 
    \left[\begin{array}{c c}
    \textbf{p}_{1} \times \textbf{p}_{2} \\
    \frac{\textbf{p}_{1} - \textbf{p}_{2}} {\left | {\textbf{p}_{1} - \textbf{p}_{2}} \right |}
    \end{array}\right] ^\top 
\end{equation}

For the remaining lines, we adopt the classical plane intersection method for spatial line initialization. Specifically, for a given line feature with $n$ observations within the sliding window, each observation corresponds to a back-projected plane in 3D space. By selecting the first observation as the reference, intersections between this reference plane and the remaining $n-1$ planes yield $n-1$ candidate 3D lines. The initial estimate of the spatial line is then obtained by averaging both the normal vectors and direction vectors of these intersected lines. The detailed derivation of this method can be found in~\cite{c13}, Algorithm B.

\subsubsection{3D Line Optimization Refinement}
After initialization, an optimization refinement step is performed to further improve the accuracy of the line triangulation. To ensure the numerical stability, we use the orthonormal representation for optimization refinement~\cite{c44}. Plücker representation has 6 DoF while a 3D line is 4 DoF, direct optimization results in over-parameterization and may degrade performance. Therefore, we adopt the orthonormal representation, which encodes the line using exactly 4 DoF, ensuring numerically stable optimization. In this representation, a spatial line is described by a 3D rotation matrix $\mathbf{U}$ and a 2D rotation matrix $\mathbf{W}$:
\begin{equation}
\mathbf{L}_{o} = (\mathbf{U},\ \mathbf{W}) \in SO(3) \times SO(2)
\end{equation}
A spatial line $\mathbf{L}_{p}$ represented in Plücker coordinates can be easily decomposed into orthonormal representation as:
\begin{equation}
\mathbf{L} = \begin{bmatrix} \mathbf{n} & \mathbf{v} \end{bmatrix} = 
\underbrace{
\begin{bmatrix} 
\dfrac{\mathbf{n}}{\|\mathbf{n}\|} & 
\dfrac{\mathbf{v}}{\|\mathbf{v}\|} & 
\dfrac{\mathbf{n} \times \mathbf{v}}{\|\mathbf{n} \times \mathbf{v}\|} 
\end{bmatrix}
}_{\mathbf{U} \in SO(3)}
\underbrace{
\begin{bmatrix} 
\|\mathbf{n}\| & 0 \\ 
0 & \|\mathbf{v}\| \\ 
0 & 0 
\end{bmatrix}
}_{\mathbf{\Sigma}_{3 \times 2}}
\end{equation}
Here, the second part is a diagonal matrix and $\mathbf{W}$ can be recovered by normalizing these nonzero components as:
\begin{equation}
\mathbf{W} = 
\frac{1}{\sqrt{ \| \mathbf{n} \|^{2} + \| \mathbf{v} \|^{2} }}
\begin{bmatrix}
\| \mathbf{n} \| & -\| \mathbf{v} \| \\
\| \mathbf{v} \| & \| \mathbf{n} \|
\end{bmatrix}
\in SO(2)
\end{equation}

Classical line optimization methods typically refine the estimated 3D line by jointly minimizing the reprojection errors of observed line features across all frames within the sliding window. However, this approach still suffers from inaccuracies under degenerate motion. To address this limitation, we incorporate two additional constraints into the optimization: the associated 3D points lying on the line and the known line direction from 2D line classification. Incorporating these constraints not only improves estimation accuracy but also effectively mitigates the impact of degenerate motions. For a line feature with $n$ image observations, $m$ associated 3D points, and a known direction, the overall cost function is:
\begin{equation}
\begin{aligned}
\underset{\mathbf{U},\ \mathbf{W}}{\arg\min} \quad & 
(\sum_{i = 1}^{n} \left\| r_{l}^{(i)}(\mathbf{U}, \mathbf{W}) \right\|^2  
+ \sum_{j = 1}^{m} \left\| r_{pl}^{(j)}(\mathbf{U}, \mathbf{W}) \right\|^2 \\
& + \left\| r_{d}(\mathbf{U}, \mathbf{W}) \right\|^2)
\end{aligned}
\end{equation}
where $r_{l}^{(i)}$ is the reprojection error of the $i$-th observation (as defined in Eq.~(\ref{eq:line_measurement})). $r_{pl}^{(j)}$ is the point-on-line error for the $j$-th 3D point, measuring the Euclidean distance between the point $\mathbf{p}$ and the spatial line $\mathbf{L}_{p}$:
\begin{equation}
d(\mathbf{p}, \mathbf{L}) = \frac{ \| \mathbf{v} \times \mathbf{p} - \mathbf{n} \| }{ \| \mathbf{v} \| }
\end{equation}
and $r_d$ is the direction alignment error between the estimated line direction $\mathbf{d}$ and the known reference direction $\hat{\mathbf{d}}$:
\begin{equation}
    r_{d} = \mathbf{d} - \hat{\mathbf{d}}
\end{equation}
The Levenberg–Marquardt algorithm is adopted to solve this nonlinear optimization problem, with a maximum of five iterations, owing to its robustness and computational efficiency.

\subsection{Implementation to Stereo Configuration}
PL-VIWO2 supports both monocular and stereo configurations. Compared with the monocular setup, which relies on a single camera, the stereo configuration is more complex as it incorporates additional information from two cameras.

For line matching in stereo configurations, after extracting line features from both the left and right images, the features are tracked through three operations: first between the left and right images, and then between these stereo pairs and the corresponding previous frame. Each matching is performed using the method described in Sec.\ref{2DProcess}. To address the additional computational cost introduced by stereo line feature detection and tracking, multi-threading is employed to parallelize the processing, thereby ensuring real-time performance.

Another difference is 3D line triangulation in the stereo setup. If a line cannot be recovered from points and directions, the plane-intersection initialization is applied, which leverages planes formed by corresponding left and right lines at the same timestep. When fewer than five stereo planes are available, additional planes are supplemented using those constructed from consecutive frames of the left camera. In the subsequent optimization refinement, observations from both cameras are jointly incorporated into the optimization process.

\subsection{Motion Consistency Check}
Urban navigation environments often contain dynamic objects, such as moving vehicles and pedestrians. To mitigate the impact of dynamic features on system performance, we introduce a motion consistency check (MCC) module that filters out inconsistent features before they are used for state updates. The core idea of MCC is that the motion of static features in 2D images should be consistent with the motion estimated by internal sensors, such as the IMU and wheel encoder. Features exhibiting large deviations are regarded as dynamic and subsequently discarded from visual update.
\begin{equation}
    \textbf{r} = \frac{1}{n} \sum \limits_{i=1}^{n}\left\| \textbf{z}_{i} - \boldsymbol{\pi} ({}_{G}^{C_{i}}\textbf{R} {} (^{G}\hat{\textbf{p}}_{f} - {}^{G}\textbf{p}_{C_{i}})) \right\| 
\end{equation}
where $\textbf{z}{i}$ is the $i$-th image measurement and $^{G}\hat{\textbf{p}}{f}$ is the triangulated 3D position of the point feature. The term $^{G}\textbf{p}{C{i}}$ denotes the camera pose stored in the sliding window. Since the IMU propagation and wheel update processes run independently of the visual update thread, the camera pose is derived from IMU and wheel measurements together with the extrinsic parameters between camera and IMU.

For point features, although a large number of 2D points are typically extracted and tracked, only a selected subset is used for state updates to maintain real-time performance. In our system, when the maximum number of static feature points for update is not yet reached, the system actively searches among all tracked points for additional candidates that pass the MCC check. For example, in our implementation, when 150 feature points are being tracked and the system is configured to use up to 70 points for state update, it does not simply select the first 70 tracked points. Instead, it iterates through all tracked points and selects the first 70 that pass the MCC. This guarantees that a sufficient number of reliable feature points are used for the state estimation, ensuring robustness in dynamic environments.

For line features, since our system only utilizes those associated with feature points, the total number of tracked line features remains relatively low — typically between 30 and 60. Therefore, similar to MINS~\cite{c4}, we apply the chi-square test to deal with dynamic lines, and only those passing the test are subsequently used for state update.

\section{Simulations}
\label{sec:simulation}
\begin{figure}[t]
    \centering
    \subfloat[Scenarios 1]{\includegraphics[width=0.23\textwidth]{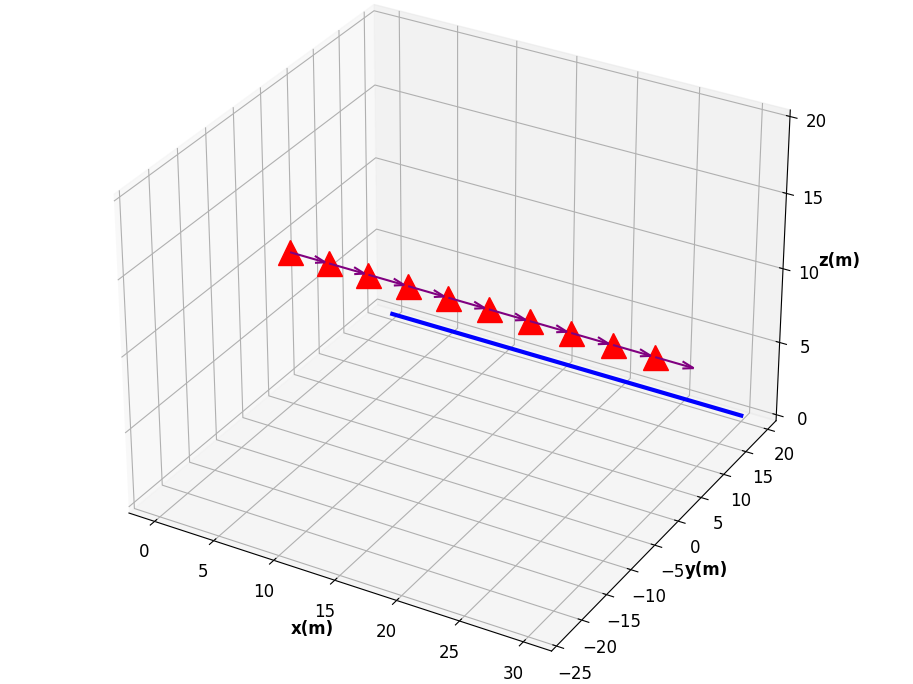}}
    \subfloat[Scenarios 2]{\includegraphics[width=0.23\textwidth]{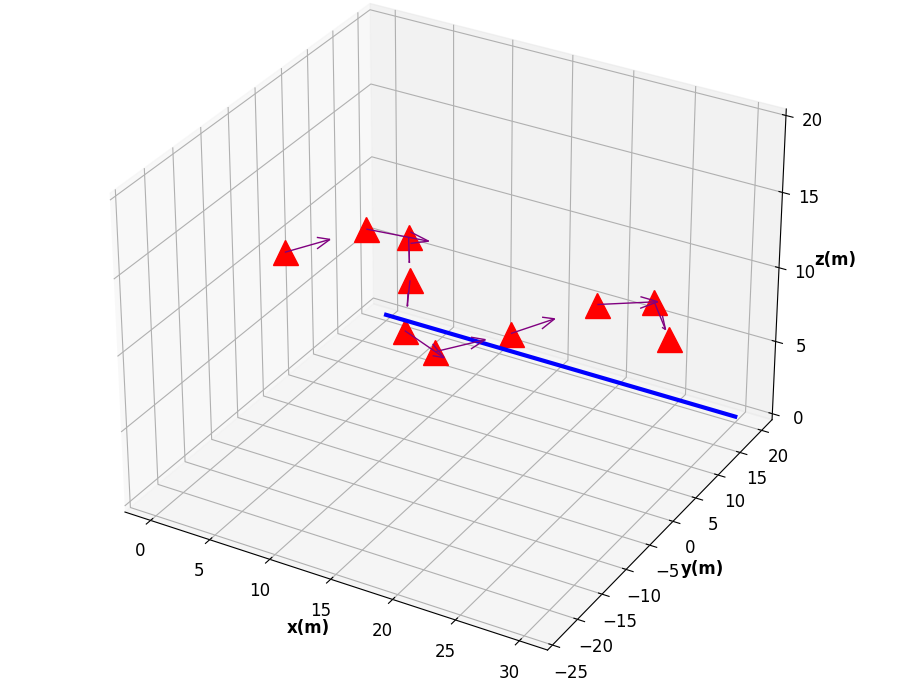}} \\
    \subfloat[Scenarios 3]{\includegraphics[width=0.23\textwidth]{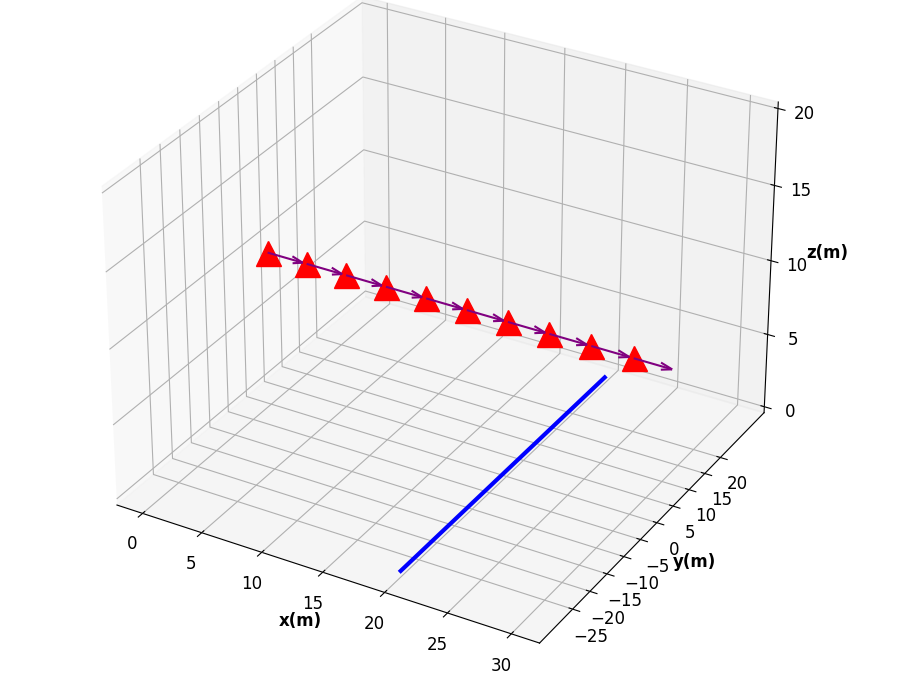}}
    \subfloat[Scenarios 4]{\includegraphics[width=0.23\textwidth]{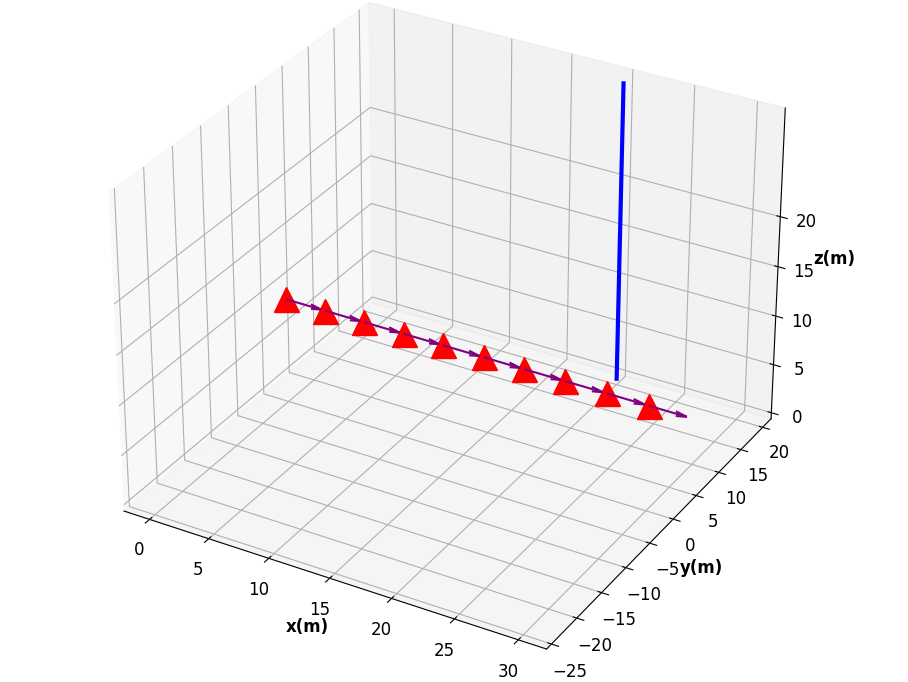}}
    \caption{Monte Carlo simulation environments for 3D line triangulation under different camera motion modes. The blue lines indicate the spatial 3D lines, red triangles represent camera positions, and the purple arrows show the camera orientation.}
    \label{fig:SimSetUp}
\end{figure}

Monte Carlo simulation experiments were conducted to validate the performance of the proposed 3D line triangulation pipeline. To evaluate the effects of different camera motions and line configurations, four representative scenarios were designed to simulate typical conditions encountered in autonomous driving, as illustrated in Fig.~\ref{fig:SimSetUp}. Three of these scenarios involve linear camera motion along straight paths in three spatial line directions respectively, simulating the straight driving conditions that vehicles commonly experience on urban roads. For example, in the first scenario, the camera moves in the same direction as the spatial line, representing a degenerate motion case used to assess its impact on the proposed pipeline. BY contrast, scenario 2 features the same spatial line but a different, non-linear camera trajectory, which is used to demonstrate that non-linear camera motion avoids degenerate conditions. In each scenario, one 3D line segments were generated and observed from ten distinct camera poses, spatially distributed throughout the scene. For each observation, the endpoints of each 3D line were projected onto the image plane, simulating the output of a line segment detector.

Triangulation accuracy is a key metric for evaluating the performance of 3D line triangulation algorithms. Since a Plücker line is represented by a normal vector and a direction vector, it is necessary to assess the accuracy of both components. To quantitatively evaluate triangulation performance, two error metrics are defined: $e_{\text{norm}}$ for the normal vector and $e_{\text{dir}}$ for the direction vector:
\begin{equation}
    e_{\text{norm}} = \|\hat{\mathbf{n}} - \mathbf{n}\|, \quad
    e_{\text{dir}} = \left\| \dfrac{\hat{\mathbf{v}}}{\|\hat{\mathbf{v}}\|} \times \dfrac{\mathbf{v}}{\|\mathbf{v}\|} \right\|.
\end{equation}
Here, $\hat{\mathbf{n}}$ and $\hat{\mathbf{v}}$ denote the estimated norm and direction vectors of the Plücker line, respectively, while $\mathbf{n}$ and $\mathbf{v}$ are the corresponding ground-truth. 

\subsection{Evaluation on Line Initialization}
First, we evaluate the performance of different 3D line initialization methods in different scenarios. Specifically, we assess three initialization strategies: (i) intersection of planes, (ii) a single point and a direction, and (iii) two distinct points. The latter two methods require additional point observations to be effective. To address this, we extract the midpoint and the 1/4 length point along the 3D segment defined by the line endpoints. These points are perturbed with noise consistent with that applied to the 2D line endpoint observations and are subsequently triangulated. For the second method, which requires a direction vector, we assume an accurate line direction can be obtained from vanishing point estimation, and thus directly utilize the ground-truth direction. 

\begin{figure*}[t]
\centering
    \subfloat{\includegraphics[width=0.45\textwidth]{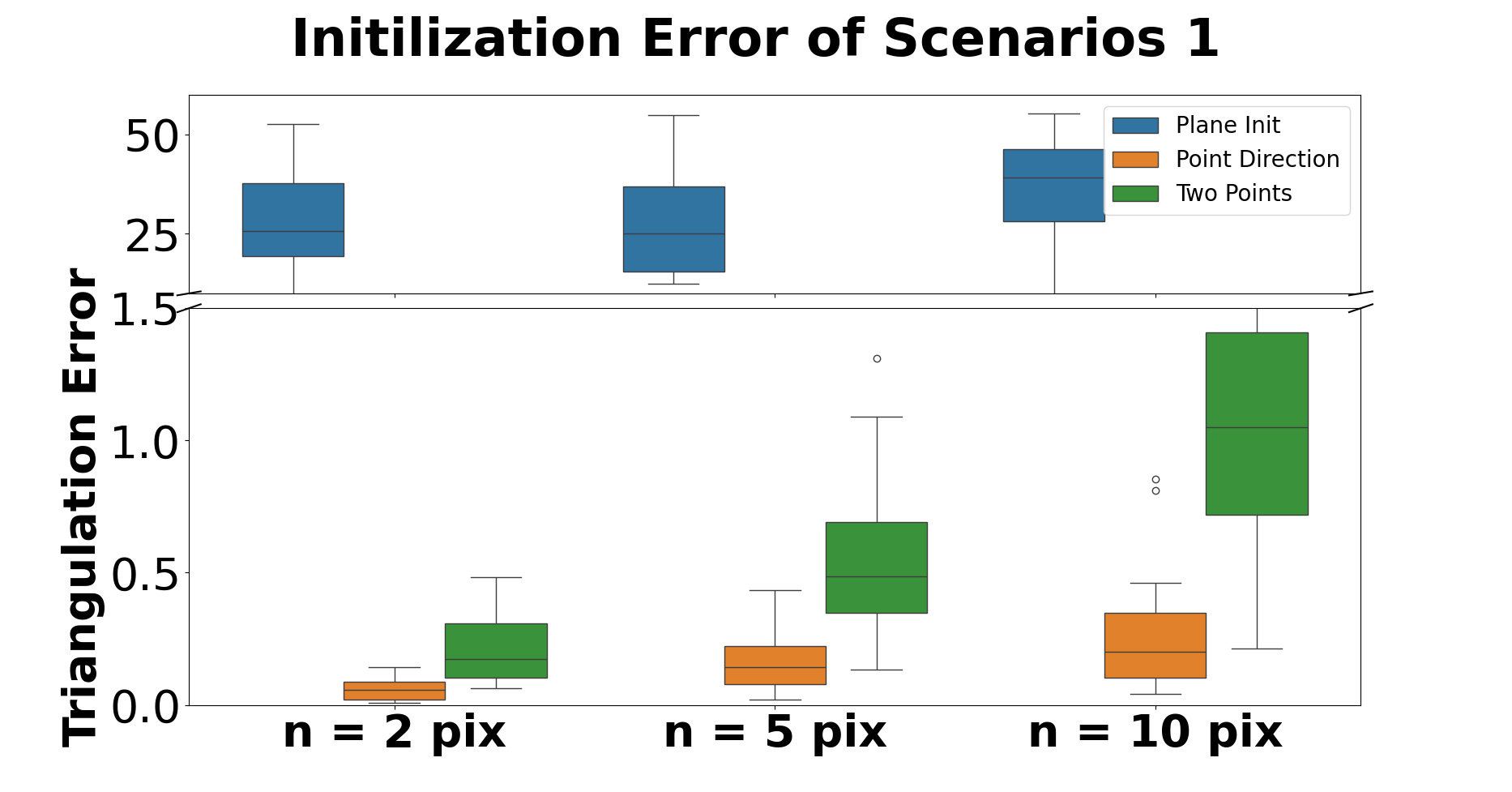}}
    \hspace{5mm}
    \subfloat{\includegraphics[width=0.45\textwidth]{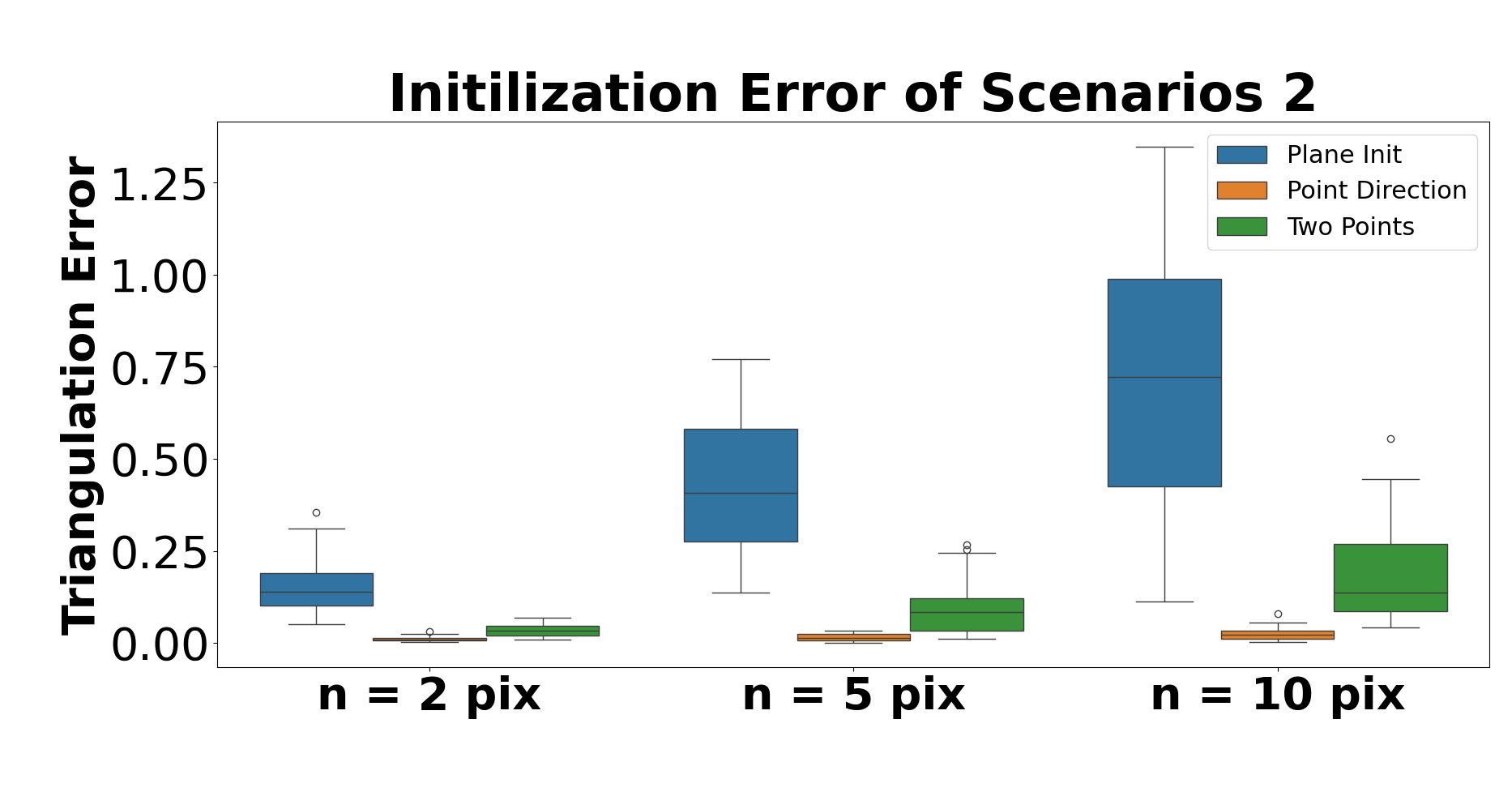}} \\
    \subfloat{\includegraphics[width=0.45\textwidth]{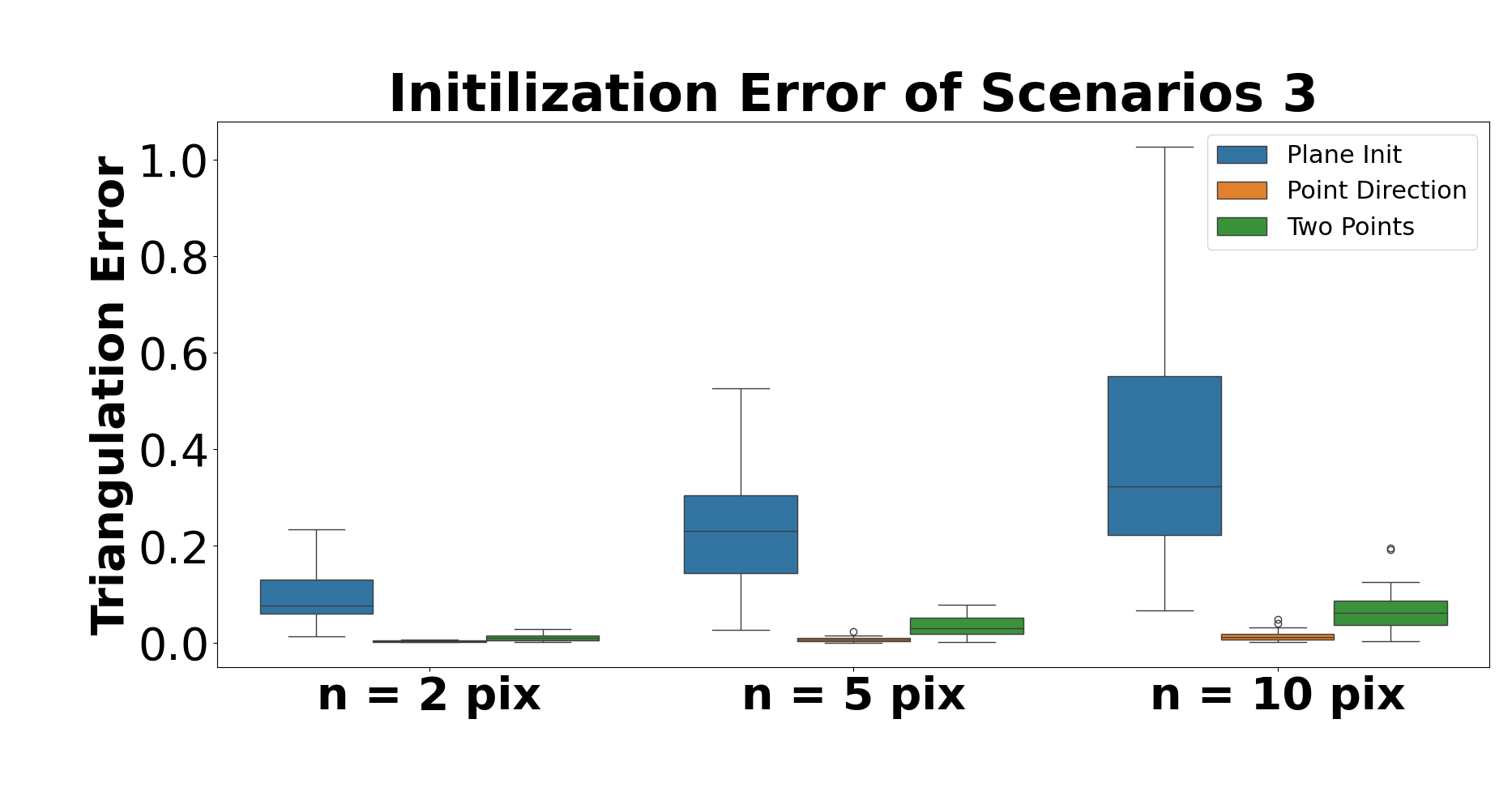}}
    \hspace{5mm}
    \subfloat{\includegraphics[width=0.45\textwidth]{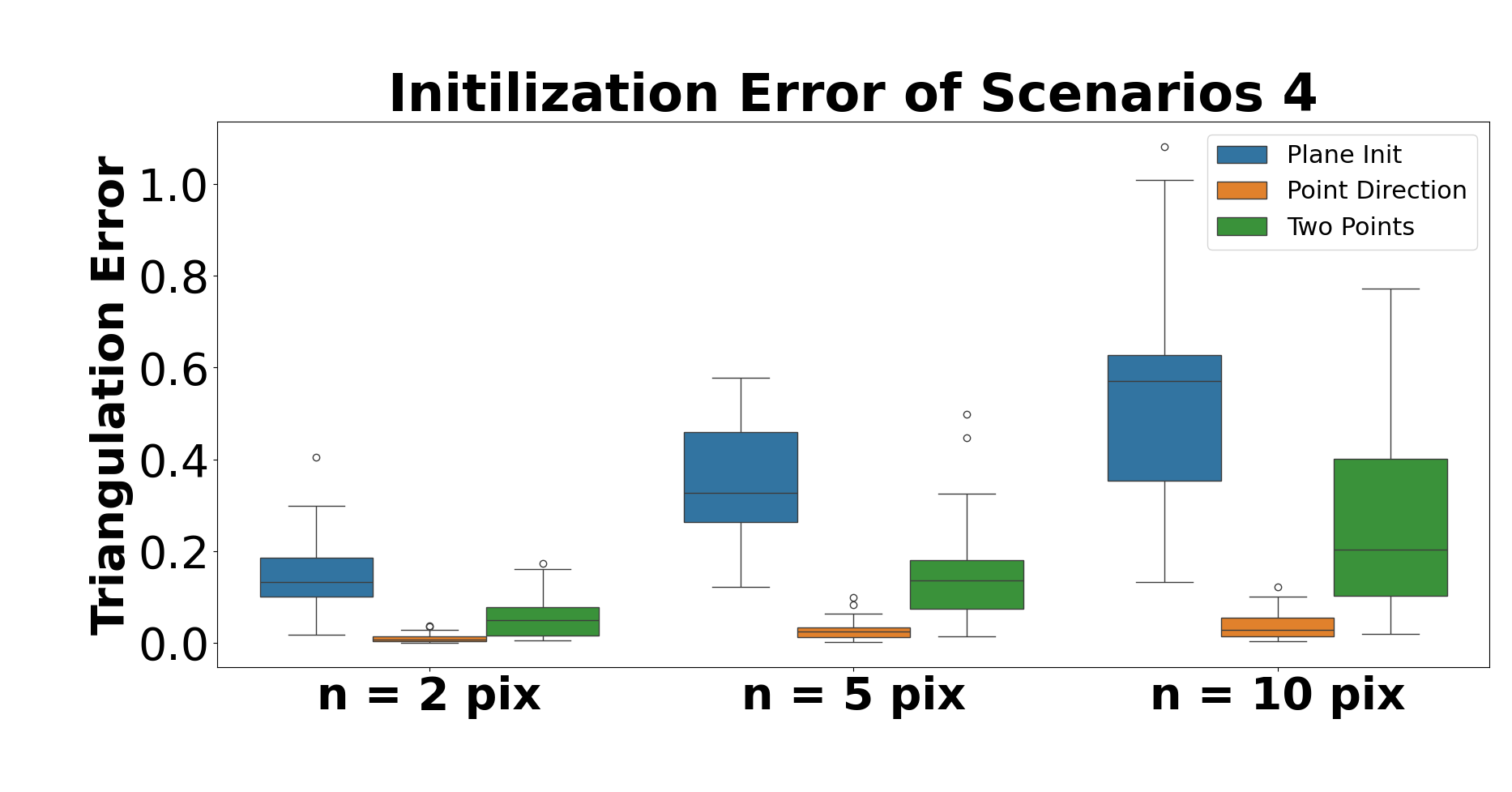}}
    \caption{Triangulation error comparison over 30 simulation tests using different initialization methods under different scenarios.}
    \label{fig:SimInitilization}
\end{figure*}

Since the direction vector is a unit vector and its error variation is relatively small, we use $e_{norm}$ as the indicator for evaluating triangulation performance under different observation noise levels. The comparative evaluation results are presented in Fig.~\ref{fig:SimInitilization}. In general, the errors of all three methods increase as the noise level rises. Among them, the method that leverages point and line direction information consistently achieves the lowest error, owing to the deterministic nature of the line direction. The approach based on two 3D points delivers moderate accuracy, while the plane intersection method typically results in the largest errors. Notably, in Scenario 1, the plane intersection method incurs significant errors during line initialization due to degenerate motion. In this case, the planes should ideally be parallel; however, observation noise causes them to intersect erroneously, leading to incorrect line estimation and large initialization errors. In contrast, the other two methods remain reliable under the same conditions.

\subsection{Evaluation on Optimization Refinement}
\label{sec:optimization}
Monte Carlo simulations were further conducted to validate the optimization algorithm based on end-point observations under different scenarios. To simulate real-world uncertainties, Gaussian noise was introduced to both the initial estimates and the camera measurements. Due to the orthogonality constraint between the direction and norm vector in the Plücker representation, noise cannot be added directly to these components. Instead, perturbations were applied by adding Gaussian Noise to the two endpoints of the 3D line segments. To simulate measurement uncertainty, observation noise $\mathbf{n}_{ob} \sim \mathcal{N}(\mathbf{0}, {\sigma}_{ob}^2\mathbf{I}_2)$ is added directly to the 2D image observations in pixel units. To validate effectiveness and convergence, the Gauss-Newton method is employed with a fixed number of 50 iterations. 

\begin{figure}
    \centering
    \subfloat{\includegraphics[width=0.45\textwidth]{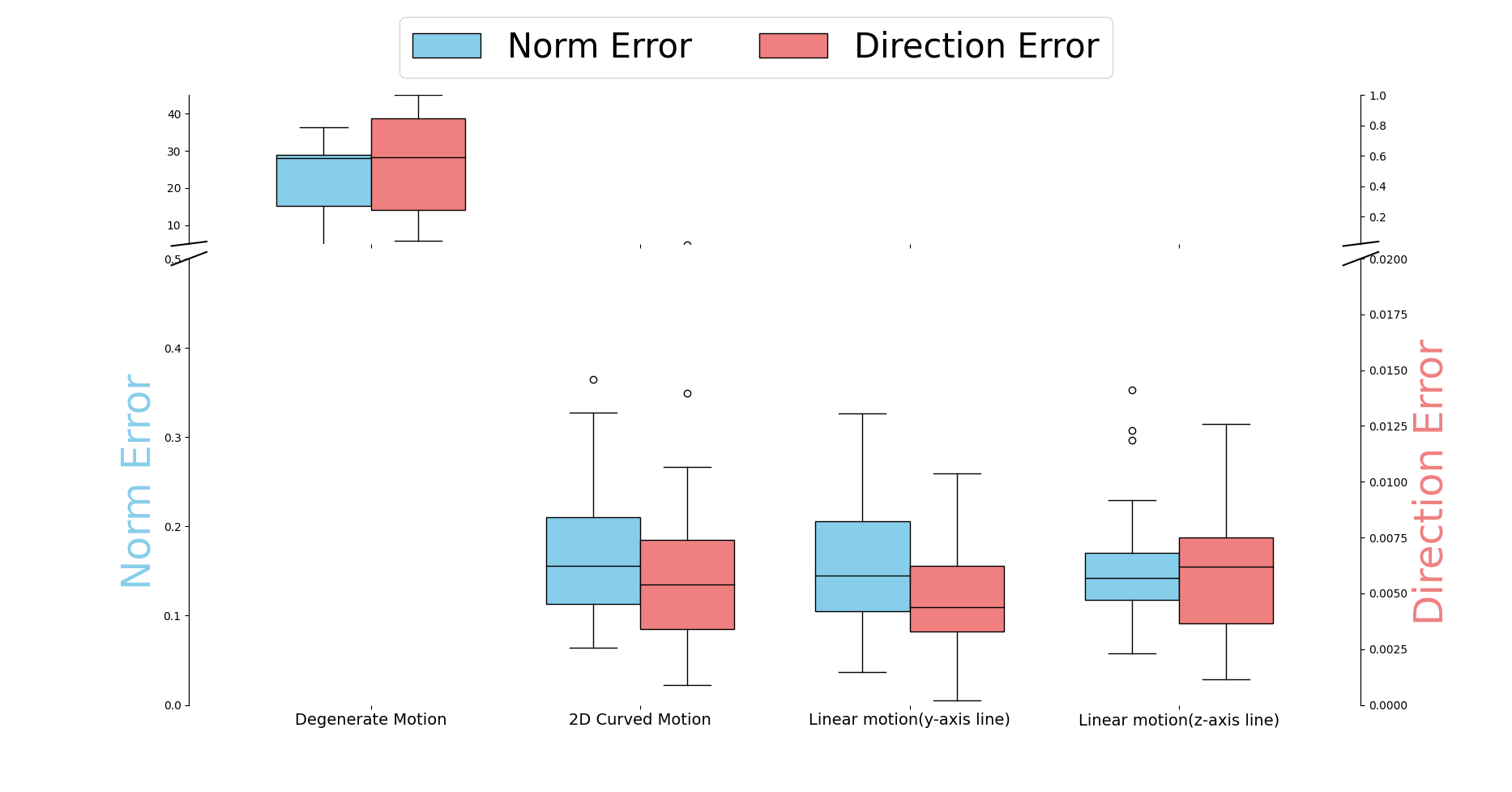}}
    \caption{Triangulation error after 50 iterations over 30 tests under different scenarios}
    \label{fig:degerate}
\end{figure}

First, to assess the impact of degenerate motion on optimization performance, 30 Monte Carlo simulations were conducted for each scenario. The resulting triangulated lines were then compared against the ground truth, as shown in Fig.\ref{fig:degerate}. In the last three scenarios, the triangulation errors are small, with the norm vector error around 0.15m and the direction vector error approximately 0.005 degrees. However, for the first scenario, which is a degenerate case, the final optimization result deviates significantly from the ground truth, with errors of about 30~m and 0.7 degrees, respectively.

We then incorporate additional information into the optimization under the same configuration, such as a 3D point on the line and the line’s spatial direction. Based on the types of information used, the resulting optimization strategies are categorized into four distinct types, as detailed below:
\begin{itemize}
    \item \textbf{Type 1:} Two endpoints;
    \item \textbf{Type 2:} Two endpoints, one 3D point on the line, and the line’s direction;
    \item \textbf{Type 3:} Two endpoints and two 3D points on the line;
    \item \textbf{Type 4:} Two endpoints, two 3D points on the line, and the line’s direction.
\end{itemize}

For the observation of endpoints, the observation and initialization noise are set the same as described in Section~\ref{sec:optimization}. For the 3D point observation, noise is added to its projected point at the same level as the line endpoints, and this noisy observation is then used to triangulate the 3D point. For the direction vector observation, the direction estimated from the vanishing points is assumed to be accurate (no noise is added). 

The triangulation results for different configurations under various test scenarios, over 30 Monte Carlo runs, are summarized in Table~\ref{tab:triangulation_error_dualrow}. In all scenarios, Type 4, which incorporates the most information, consistently achieves the lowest triangulation errors. Furthermore, when direction observations are introduced in Type 2 and Type 4, the direction vector error is minimized because the observed direction is assumed to be the ground truth. It is also noteworthy that in the first scenario, due to the degenerate motion pattern, the error for Type 1 is significantly higher than that of the other three types. However, after introducing additional 3D point and direction vector information, the optimization error is substantially reduced.

\begin{table*}
    \centering
    \caption{Mean and standard deviation of Triangulation errors for each type under different scenarios.}
    \setlength{\tabcolsep}{15pt}
    \resizebox{\textwidth}{!}{
    \begin{tabular}{c|cccc}
        \hline
        Scenario & Type 1 & Type 2 & Type 3 & Type 4 \\
        \hline
        \multirow{2}{*}{Scenario 1} 
        & \textcolor{red}{25.3068 ($\pm$ 5.6271)} & 0.1074 ($\pm$ 0.0497) & 0.1091 ($\pm$ 0.0480) & \textbf{0.1051 ($\pm$ 0.0478)} \\
         & \textcolor{red}{0.7058 ($\pm$ 0.2904)} & \textbf{0.0036 ($\pm$ 0.0017)} & 0.0037 ($\pm$ 0.0017) & \textbf{0.0036 ($\pm$ 0.0017)} \\
        \hline
        \multirow{2}{*}{Scenario 2} 
        & 0.2011 ($\pm$ 0.0957) & 0.0792 ($\pm$ 0.0379) & 0.0790 ($\pm$ 0.0377) & \textbf{0.0787 ($\pm$ 0.0368)} \\
        & 0.0077 ($\pm$ 0.0033) & \textbf{0.0030 ($\pm$ 0.0013)} & 0.0032 ($\pm$ 0.0014) & \textbf{0.0030 ($\pm$ 0.0013)} \\
        \hline
        \multirow{2}{*}{Scenario 3} 
        & 0.1994 ($\pm$ 0.0751) & 0.0658 ($\pm$ 0.0285) & 0.0777 ($\pm$ 0.0295) & \textbf{0.0650 ($\pm$ 0.0279)} \\
        & 0.0073 ($\pm$ 0.0031) & \textbf{0.0026 ($\pm$ 0.0012)} & 0.0029 ($\pm$ 0.0013) & \textbf{0.0026 ($\pm$ 0.0012)} \\
        \hline
        \multirow{2}{*}{Scenario 4} 
        & 0.2069 ($\pm$ 0.0836) & 0.0736 ($\pm$ 0.0331) & 0.0797 ($\pm$ 0.0333) & \textbf{0.0725 ($\pm$ 0.0331)} \\
        & 0.0067 ($\pm$ 0.0028) & \textbf{0.0025 ($\pm$ 0.0011)} & 0.0027 ($\pm$ 0.0011) & \textbf{0.0025 ($\pm$ 0.0011)} \\
        \hline
    \end{tabular}
    }
    \label{tab:triangulation_error_dualrow}
\end{table*}


\section{Real World Experiments}
\label{sec:experiments}
To evaluate the real-world performance of PL-VIWO2, we conducted experiments on the public KAIST Urban Dataset\cite{c47}. This dataset was collected using a vehicle equipped with a PointGrey Flea3 stereo camera (10 Hz), an Xsense MTi-300 MEMS IMU (100 Hz), and an RLS LM13 photoelectric wheel encoder (100 Hz). Based on the driving scenarios, the dataset can be broadly categorized into highway and urban road environments, with representative samples shown in Fig.~\ref{fig:RepresentativeImages}. Each sequence presents a wide range of challenging scenarios, including dynamic objects, overexposure, texture-less regions, and long-term driving. These diverse and complex conditions make the KAIST Urban Dataset a comprehensive benchmark for evaluating system performance in highly complex urban environments. The benchmark methods are as follows:
\begin{figure}[t]
\centering
    \subfloat[Highway Scenarios]{\includegraphics[width=0.22\textwidth]{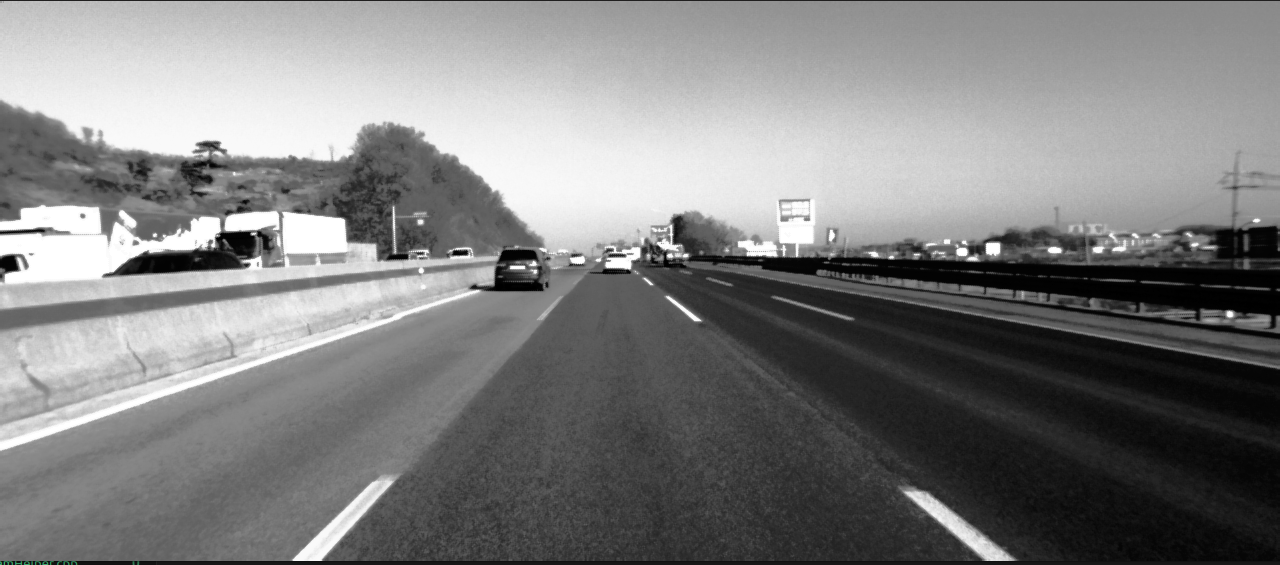}}
    \hspace{0.01\textwidth}
    \subfloat[Urban Scenarios]{\includegraphics[width=0.22\textwidth]{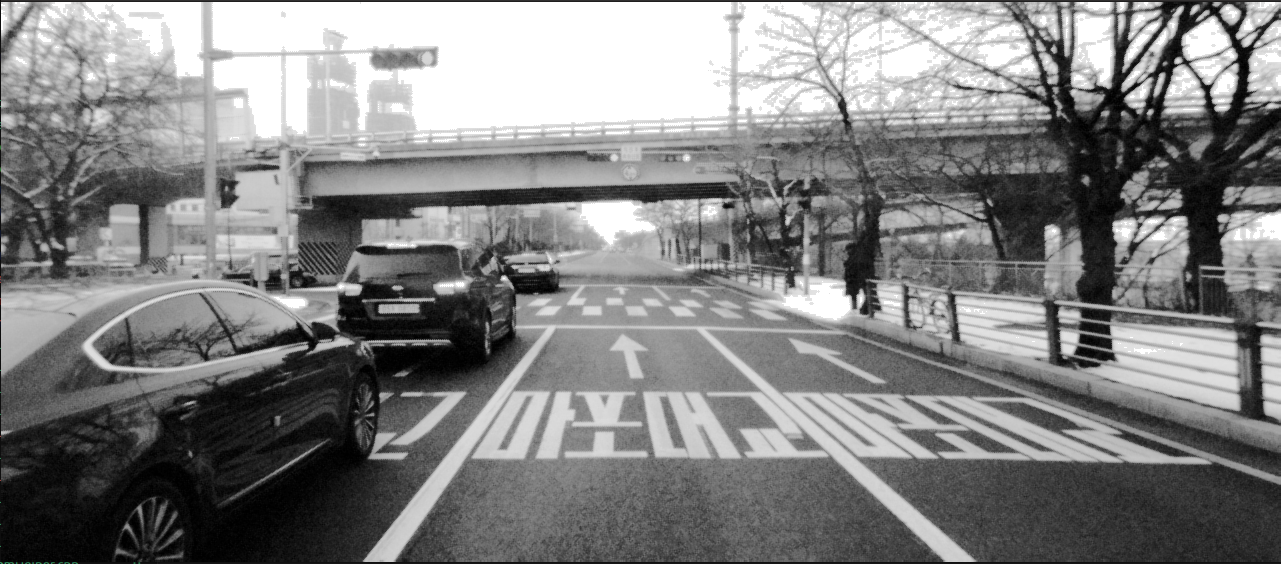}}
    \caption{Representative images from the KAIST Urban Datasets.}
    \label{fig:RepresentativeImages}
\end{figure}
\begin{figure}
    \centering
    \includegraphics[width=0.9\linewidth]{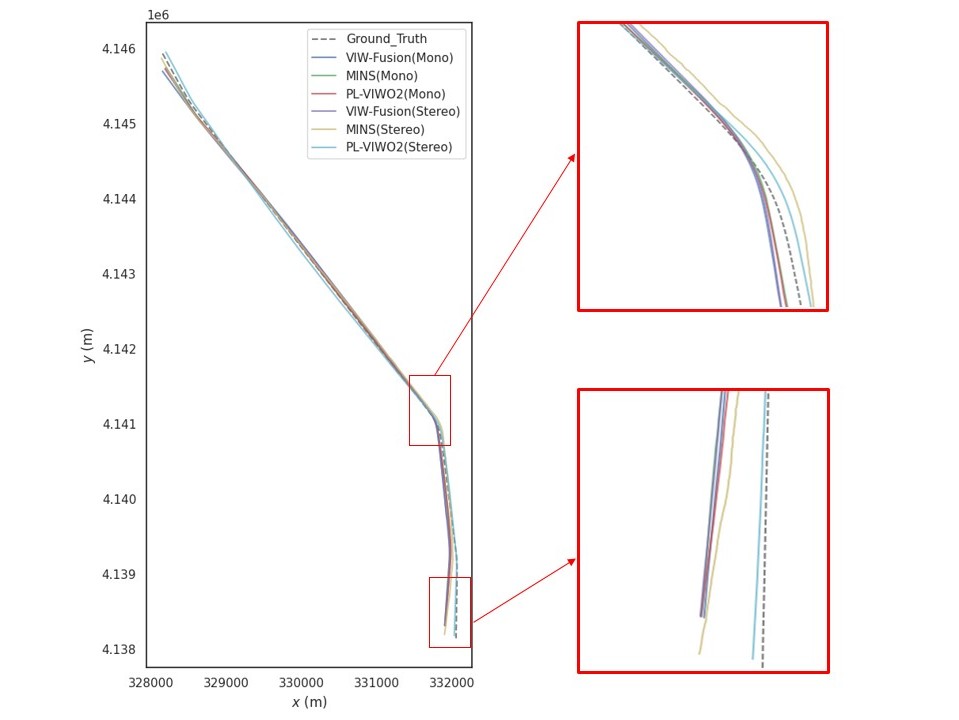}
    \caption{Trajectory comparison of multiple algorithms on KAIST Urban36.}
    \label{fig:urban36}
\end{figure}
\begin{table}[htbp]
  \centering
  \caption{Line Tracking Performance Across Two Driving Scenarios}
  \begin{tabular}{c lcc}
    \hline
     & \textbf{Method} & \textbf{Matching Rate (\%)} & \textbf{Processing Time (ms)} \\
    \hline
    \multirow{4}{*}{\rotatebox{90}{Urban20}} 
        & Method 1 & 44.58 & 0.3 \\
        & Method 2 & 43.37 & 0.5 \\
        & Method 1+2       & 58.42 & 0.6 \\
        & LBD                  & 16.67 & 5.0 \\
    \hline
    \multirow{4}{*}{\rotatebox{90}{Urban38}} 
        & Method 1 & 71.50 & 0.5 \\
        & Method 2 & 73.63 & 0.7 \\
        & Method 1+2       & 82.68 & 0.9 \\
        & LBD                  & 31.36 & 6.0 \\
    \hline
  \end{tabular}
  \label{tab:line_tracking_results}
\end{table}
\begin{figure}
    \centering
    \includegraphics[width=1.0\linewidth]{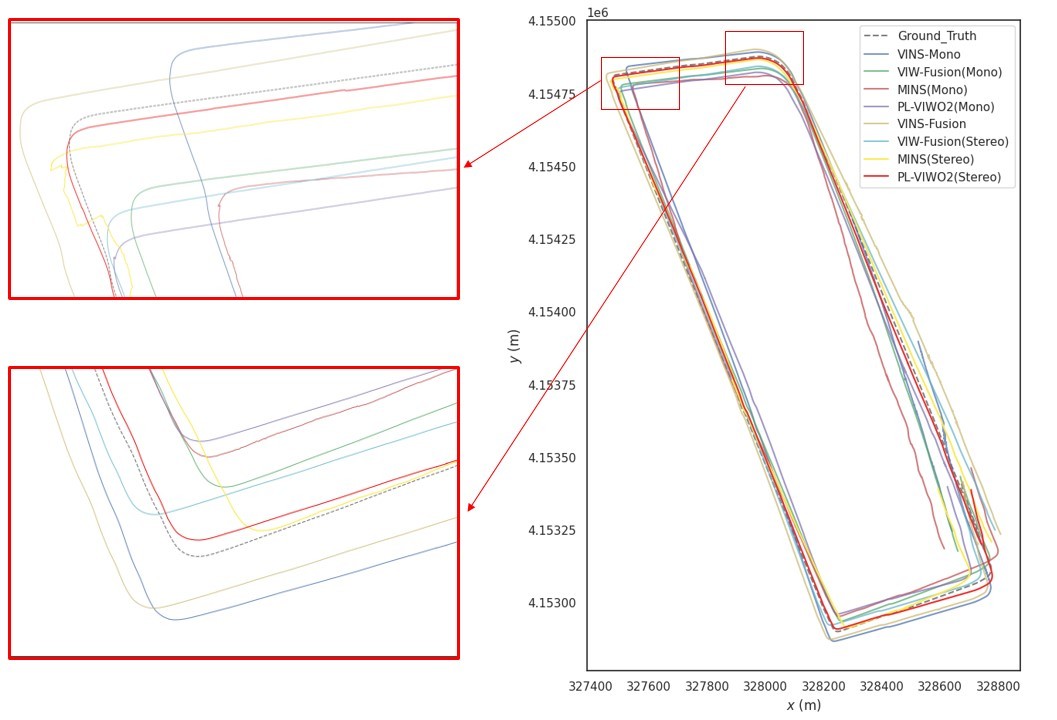}
    \caption{Trajectory comparison of multiple algorithms on KAIST Urban30.}
    \label{fig:urban30}
\end{figure}
\begin{itemize}
    \item \textbf{VINS-Mono}~\cite{c2}: A classical monocular VIO system that tightly couples IMU and camera measurements within a factor graph optimization framework.
    \item \textbf{VINS-Fusion}~\cite{c48}: An extension of VINS-Mono that supports stereo + IMU, and stereo-only configurations. We use the stereo + IMU mode in our experiments as a benchmark for stereo-based VIO systems.
    \item \textbf{PL-VINS}~\cite{c27}: An extension of VINS-Mono that integrates line features. Line matching is performed using descriptor-based methods, which serves as a baseline for evaluating proposed line matching approach.
    \item \textbf{VIW-Fusion}~\cite{c49}: A VIWO system built upon VINS-Fusion that incorporates wheel odometry measurements and enforces planar motion constraints.
    \item \textbf{OpenVINS}~\cite{c3}: A MSCKF-based VIO system that supports both monocular and stereo configurations.
    \item \textbf{MINS}~\cite{c4}: A multi-sensor fusion odometry framework based on MSCKF, which supports various sensor modalities. Since our system is developed on top of this framework, MINS serves as a key benchmark for evaluating our contributions. For fair comparison, we adopt its configuration that fuses camera, IMU, and wheel odometry as a baseline for VIWO evaluation.
\end{itemize}

To validate the proposed methods, we first evaluate the accuracy and efficiency of line matching using various techniques. We then compare the localization performance of PL-VIWO2 in both monocular and stereo modes against state-of-the-art benchmarks. An ablation study is conducted to assess the contribution of each individual module to the overall system accuracy. In addition, the processing time of each module is recorded, and the overall computational efficiency is analyzed in comparison with representative benchmarks. Finally, a set of line triangulation results is presented to further demonstrate the robustness and precision of the proposed method. All experiments are conducted on a personal computer equipped with an Intel Core i7 CPU, 32 GB of RAM, and an NVIDIA A1000 GPU (6 GB VRAM), running Ubuntu 20.04.
\begin{table*}
\centering
\renewcommand\arraystretch{1.0}
\caption{RMSE ATE for Highway Sequences: Upper row is Translation Error (m), lower row is Rotation Error (°).}
\label{table1}
\setlength{\tabcolsep}{0.5pt}
\begin{tabular}{|m{3.6cm}<{\centering}|m{1.2cm}<{\centering}|m{1.2cm}<{\centering}|m{1.2cm}<{\centering}|m{1.2cm}<{\centering}|m{1.2cm}<{\centering}|m{1.2cm}<{\centering}|m{1.2cm}<{\centering}|m{1.2cm}<{\centering}|m{1.2cm}<{\centering}|m{1.2cm}<{\centering}|m{1.2cm}<{\centering}|} 
\hline 
Algorithms & urban18 (3.9km) & urban19 (3.0km) & urban20 (3.2km) & urban21 (3.7km) & urban22 (3.4km) & urban23 (3.4km) & urban24 (4.2km) & urban25 (2.5km) &urban35 (3.2km) & urban36 (9.0km) & urban37 (11.8km) \\  \hline
\multicolumn{12}{|c|}{\textbf{Monocular VIO Methods}} \\ \hline
\rowcolor{gray!30} {VINS-mono~\cite{c2}} & {FAIL} & {FAIL} & {FAIL} & {FAIL} & {FAIL} & {FAIL} & {FAIL} & {FAIL} & {FAIL} & {FAIL} & {FAIL} \\
\rowcolor{gray!30} & {} & {} & {} & {} & {} & {} & {} & {} &  & {} & {} \\ \hline

\rowcolor{gray!30} {PL-VINS~\cite{c27}} & {FAIL} & {FAIL} & {FAIL} & {FAIL} & {FAIL} & 208.26 & {FAIL} & {FAIL} & 256.70 & {FAIL} & {FAIL} \\
\rowcolor{gray!30}   & &  &  &  &  & 7.10 &  &  & 159.09 &  &  \\ \hline  
\rowcolor{gray!30} {VIW-Fusion~\cite{c49}} & \textbf{43.60} & 37.57 & \underline{38.65} & 52.86 & 53.05 & \underline{42.28} & 55.88 & \textbf{29.46} & 36.77 & \underline{116.98} & \underline{174.87} \\
\rowcolor{gray!30}    & 4.27 & 6.80 & 155.40 & 160.25 & 156.00 & 12.00 & 55.12 & \underline{9.01} & \underline{10.35} & 20.02 & 158.55 \\ \hline
{OpenVINS~\cite{c3}} & {FAIL} & {FAIL} & {FAIL} & {FAIL} & {FAIL} & {FAIL} & {FAIL} & {FAIL} & {FAIL} & {FAIL} & {FAIL} \\
-mono &  &  &  &  &  &  & & &  &  &  \\ \hline
{MINS(I,C,W)~\cite{c4}} & 50.58 & \underline{37.19} & 40.67 & \underline{46.33} & \underline{46.12} & 42.53 & \textbf{47.65} & 36.60 & \underline{55.01} & 117.89 & 197.53 \\
-mono & \underline{2.94} & \underline{6.02} & \underline{9.39} & \underline{12.67} & \underline{16.71} & \textbf{5.95} & \underline{36.93} & 37.30 & 15.68 & \underline{6.34} & \underline{24.16} \\ \hline    
{PL-VIWO2} & \underline{47.99} & \textbf{35.74} & \textbf{38.51} & \textbf{46.05} & \textbf{44.01} & \textbf{40.36} & \underline{51.77} & \underline{31.56} & \textbf{35.38} & \textbf{105.46} & \textbf{159.36} \\
-mono & \textbf{2.80} & \textbf{4.46} & \textbf{3.59} & \textbf{4.36} & \textbf{15.90} & \underline{9.22} & \textbf{23.35} & \textbf{3.93} & \textbf{3.97} & \textbf{6.26} & \textbf{13.52} \\ \hline
\multicolumn{12}{|c|}{\textbf{Stereo VIO Methods}} \\ \hline
\rowcolor{gray!30} {VINS-Fusion\cite{c48}} & {FAIL} & {FAIL} & {FAIL} & {FAIL} & {FAIL} & {FAIL} & {FAIL} & {FAIL} & {FAIL} & {FAIL} & {FAIL} \\
\rowcolor{gray!30} -stereo &  &  &  &  &  &  & & &  &  &  \\ \hline
\rowcolor{gray!30} {VIW-Fusion\cite{c49}} & 39.84 & 36.68 & 40.06 & 44.68 & 55.29 & 44.97 & 51.37 & 33.70 & 36.41 & 119.86 & 148.04 \\
\rowcolor{gray!30} -stereo & 6.32 & 9.80 & 98.87 & 23.05 & 35.98 & 30.17 & 45.35 & \underline{17.41} & 21.19 & 15.97 & 126.06 \\ \hline
{OpenVINS~\cite{c3}} & {FAIL} & {FAIL} & {FAIL} & {FAIL} & {FAIL} & {FAIL} & {FAIL} & {FAIL} & {FAIL} & {FAIL} & {FAIL} \\
-stereo &  &  &  &  &  &  & & &  &  &  \\ \hline
{MINS(I,C,W)~\cite{c4}} & \underline{24.40} & \textbf{28.03} & \underline{21.13} & \underline{29.60} & \underline{10.06} & \underline{30.29} & \underline{25.44} & \underline{25.70} & \underline{26.90} & \underline{57.29} & \underline{119.48} \\
-stereo & \underline{4.50} & \underline{7.86} & \underline{32.92} & \underline{4.94} & \underline{2.85} & \underline{19.08} & \underline{11.99} & 36.56 & \underline{11.23} & \underline{10.10} & \underline{16.92} \\ \hline
{PL-VIWO2} & \textbf{24.15} & \underline{30.05} & \textbf{16.92} & \textbf{18.04} & \textbf{7.19} & \textbf{28.82} & \textbf{22.09} & \textbf{19.74} & \textbf{14.33} & \textbf{34.55} & \textbf{102.05} \\
-stereo & \textbf{3.62} & \textbf{3.39} & \textbf{1.72} & \textbf{3.71} & \textbf{2.80} & \textbf{7.44} & \textbf{6.55} & \textbf{5.18} & \textbf{2.04} & \textbf{7.26} & \textbf{6.70} \\ \hline
\end{tabular}
\label{tab:Highway}
\end{table*}
\subsection{Line Matching Evaluation}
To demonstrate the efficiency of the proposed line matching methods in complex outdoor environments, a comparative evaluation is conducted on the two optical–flow–based line tracking algorithms, both individually and in combination with the traditional descriptor–based method (LBD~\cite{c26}) implemented in PL-VINS~\cite{c27}. The evaluation focuses on tracking rate and matching time between adjacent frames. To ensure a comprehensive assessment, two representative datasets are selected: Urban20 and Urban38, representing highway and urban driving respectively. It is worth noting that in PL-VINS, all line features with successfully computed descriptors are tracked, whereas in PL-VIWO2, only line features associated with point features are considered, as these can be tracked efficiently and provide richer information.

The average matching rate and processing time are summarized in Table~\ref{tab:line_tracking_results}. Method 1 achieves the lowest processing time, as it directly utilizes point feature tracking results and does not require additional computations for optical flow calculation. Method 2 attains a similar matching rate but incurs higher processing time because it tracks five points using optical flow. The combination method, which first applies Method 1 for initial tracking and then uses Method 2 for the remaining line features, achieves the highest matching rate, with only a modest increase in processing time.

In contrast, descriptor-based methods rely on line segment attributes such as orientation and length across matching frames. These attributes can vary substantially during rapid outdoor motion, leading to increased mismatches. This effect is particularly evident in highway scenarios, where the large inter-frame displacement caused by high-speed movement significantly reduces tracking reliability. Furthermore, line features located on dynamic objects may exhibit substantial descriptor variations over time, further degrading matching performance. Consequently, LBD matching not only yields a lower matching rate but also requires considerable computation time for line feature descriptor calculation.

\subsection{Localization Accuracy}
Then we evaluate the localization accuracy of PL-VIWO2 on the KAIST Urban Dataset and compare its performance with several SOTA methods. Since PL-VIWO2 supports both monocular and stereo configurations, we ensure a fair comparison by aligning the evaluation modes: the monocular version of PL-VIWO2 is compared against monocular VIO/VIWO systems, while the stereo version is evaluated against stereo systems. Due to PL-VINS only supporting monocular version, we put the monocular result only. Based on the testing scenarios, the results are categorized into highway and city environments, and are presented separately.

To ensure a fair comparison, we disabled loop closure and online calibration, including intrinsic, extrinsic parameters and time offset for all evaluated methods. The localization accuracy is assessed using the root mean squared error (RMSE) of the SE(3) absolute trajectory error (ATE), reported separately as translation error and rotation error. All evaluation is performed with the EVO toolkit~\cite{c50}. In the result tables, optimization-based algorithms are highlighted with a grey background. Methods that fail to complete the test or produce a translation error greater than 300 meters are marked as \textbf{FAIL}. For each configuration, the best results are shown in \textbf{bold}, and the second-best results are \underline{underlined}.
\begin{table*}
\centering
\renewcommand\arraystretch{1.0}
\caption{RMSE ATE for City Sequences: Upper row is Translation Error (m), lower row is Rotation Error (°).}
\label{tab:Urban}
\setlength{\tabcolsep}{1pt}
\begin{tabular}{|m{3.6cm}<{\centering}|m{1.3cm}<{\centering}|m{1.3cm}<{\centering}|m{1.3cm}<{\centering}|m{1.3cm}<{\centering}|m{1.3cm}<{\centering}|m{1.3cm}<{\centering}|m{1.3cm}<{\centering}|m{1.3cm}<{\centering}|m{1.3cm}<{\centering}|m{1.3cm}<{\centering}|}
\hline 
Algorithms & urban26 (4.0km) & urban27 (5.4km) & urban28 (11.5km) & urban29 (3.6km) & urban30 (6.0km) & urban31 (11.4km) & urban32 (7.1km) & urban33 (7.6km) &urban34 (7.8km) & urban38 (11.4km)\\      \hline
\multicolumn{11}{|c|}{\textbf{Monocular VIO Methods}} \\ \hline
\rowcolor{gray!30} {VINS-mono~\cite{c2}} & 35.43 & 151.34 & 94.77 & FAIL & 110.40 & FAIL & \underline{74.24} & FAIL & FAIL & 170.39\\ 
\rowcolor{gray!30} & 3.58 & 7.04 & \textbf{3.58} &  & 7.35 &  & \underline{5.90} &  &  & 8.40 \\ \hline 
\rowcolor{gray!30} {PL-VINS~\cite{c27}} & FAIL & FAIL & FAIL & FAIL & FAIL & FAIL & 154.41 & 134.07 & 291.33 & 284.88 \\
\rowcolor{gray!30} &  &   &  &  &   & & 8.83 & \underline{7.38} & 18.47 & 18.03  \\ \hline    
\rowcolor{gray!30} {VIW-Fusion~\cite{c49}} & \textbf{23.86} & 124.01 & \textbf{32.28} & \textbf{42.26} & \textbf{40.72} & \underline{230.15} & 86.11 & 127.52 & \underline{39.18} & \textbf{48.12}  \\
\rowcolor{gray!30} -mono & \underline{3.31} & 32.35 & \underline{5.05} & \textbf{4.70} & \textbf{5.56} & \underline{14.08} & 7.03 & 16.63 & \textbf{3.62} & \underline{7.60} \\ \hline
{OpenVINS~\cite{c3}} & {FAIL} & {FAIL} & {FAIL} & 82.41 & {FAIL} & {FAIL} & {FAIL} & {FAIL} & {FAIL} & {FAIL}\\
-mono &  &  &  & \underline{4.51} &  &  &  &  &  &  \\ \hline
{MINS(I,C,W)~\cite{c4}} & 34.87 & \underline{68.87} & 91.00 & 68.41 & 73.56 & 334.15 & 93.89 & \underline{85.47} & 40.64 & 104.04\\
-mono & 3.87 & \underline{6.79} & 11.79 & 9.80 & 10.37 & 16.83 & 9.45 & 8.20 & 6.00 & 10.78  \\ \hline
{PL-VIWO2} & \underline{27.12} & \textbf{44.81} & \underline{49.31} & \underline{45.78} & \underline{57.46} & \textbf{178.76} & \textbf{66.71} & \textbf{55.56} & \textbf{31.76} & \underline{55.31} \\
-mono & \textbf{2.40} & \textbf{2.81} & 6.52 & 6.31 & \underline{6.92} & \textbf{9.36} & \textbf{5.82} & \textbf{6.33} & \underline{5.42} & \textbf{5.69} \\ \hline
\multicolumn{11}{|c|}{\textbf{Stereo VIO Methods}} \\ \hline
\rowcolor{gray!30} {VINS-Fusion~\cite{c48}} & 44.58 & 74.50 & 59.23 & 65.72 & 45.80 & {FAIL} & {FAIL} & 42.54 & {FAIL}  &  66.43 \\
\rowcolor{gray!30} -stereo & 7.53 & 13.84 & 8.50 & \textbf{2.12} & 6.94 &  &  & 8.35 &  & 10.59 \\ \hline
\rowcolor{gray!30} {VIW-Fusion~\cite{c49}} & 22.59 & 29.31 & 40.71 & 35.73 & 38.18 & 353.58 & 78.85 & 45.22 & 47.79 & 59.62 \\
\rowcolor{gray!30} -stereo & 3.23 & 6.28 & 4.16 & 2.83 & 7.35 & 18.96 & 8.90 & 6.21 & \underline{4.58} & 6.48 \\ \hline
{OpenVINS~\cite{c3}} & \underline{5.36} & {FAIL} & 33.37 & 27.23 & {FAIL} & {FAIL} & \underline{27.80} & \underline{23.61} & {FAIL} & {FAIL} \\
-stereo & \underline{2.28} &  & 6.68 & \underline{2.49} &  &  & \underline{3.54} & \textbf{2.82} &  &  \\ \hline
{MINS(I,C,W)~\cite{c4}} & 8.17 & \underline{16.31} & \textbf{17.74} & \underline{17.17} & 33.62 & \underline{101.58} & \textbf{25.61} & 30.72 & \textbf{33.81} & \underline{15.99} \\
-stereo & 2.24 & \underline{2.76} & \underline{3.03} & 3.22 & \underline{4.19} & \underline{8.82} & 4.85 & 4.10 & 5.24 & \underline{3.16} \\ \hline
{PL-VIWO2} & \textbf{5.21} & \textbf{16.30} & \underline{19.54} & \textbf{16.85} & \textbf{14.68} & \textbf{60.98} & 30.95 & \textbf{21.96} & \underline{34.84} & \textbf{11.00} \\
-stereo & \textbf{1.89} & \textbf{2.31} & \textbf{2.69} & 3.20 & \textbf{4.01} & \textbf{4.75} & \textbf{3.41} & \underline{3.50} & \textbf{4.38} & \textbf{2.08} \\ \hline
\end{tabular}
\end{table*}

\begin{table*}
\centering
\renewcommand\arraystretch{1.0}
\caption{Ablation study of the monocular setup in city sequences. \\ 
RMSE ATE is reported for PL-VIWO, PL-VIWO2 (w/o line), PL-VIWO2 (w/o line refinement), and PL-VIWO2.}
\label{tab:ablation}
\setlength{\tabcolsep}{1pt}
\begin{tabular}{|m{3.6cm}<{\centering}|m{1.3cm}<{\centering}|m{1.3cm}<{\centering}|m{1.3cm}<{\centering}|m{1.3cm}<{\centering}|m{1.3cm}<{\centering}|m{1.3cm}<{\centering}|m{1.3cm}<{\centering}|m{1.3cm}<{\centering}|m{1.3cm}<{\centering}|m{1.3cm}<{\centering}|}
\hline 
Algorithms & urban26 & urban27 & urban28 & urban29 & urban30 & urban31 & urban32 & urban33 &urban34 & urban38\\ \hline
{MINS(I,C,W)~\cite{c4}} & 34.87 & 68.87 & 91.00 & 68.41 & 73.56 & 334.15 & 93.89 & 85.47 & 40.64 & 104.04\\
 & 3.87 & 6.79 & 11.79 & 9.80 & 10.37 & 16.83 & 9.45 & 8.20 & 6.00 & 10.78  \\ \hline
 
{PL-VIWO}\cite{c16} & 28.84 & 47.67 & 49.19 & 47.34 & 65.67 & 191.77 & 65.96 & 86.63 & 33.60 & 73.31  \\
        & \textbf{2.27} & 3.03 & 7.00 & 6.83 & \textbf{6.63} & 11.15 & 7.78 & 6.88 & 5.54 & 6.98 \\ \hline

{PL-VIWO2} & 36.30 & 55.24 & 73.54 & 66.10 & 64.01 & 302.19 & 78.08 & 75.29 & 37.99 & 96.16\\
(w/o line) & 3.51 & 7.19 & 7.70 & 9.62 & 5.54 & 13.47 & 8.32 & 9.05 & 6.19 & 9.75 \\ \hline

{PL-VIWO2} & 27.74 & 46.81 & \textbf{49.11} & 46.82 & 60.84 & 179.58 & \textbf{64.24} & 64.72 & 33.26 & 65.29   \\
(w/o line refinement)  &2.32 & 2.97 & 6.89 & 6.58 & 6.82 & 10.92 & 5.82 & 7.36 & 6.54 & 7.36 \\ \hline

{PL-VIWO2} & \textbf{27.12} & \textbf{44.81} & 49.31 & \textbf{45.78} & \textbf{57.46} & \textbf{178.76} & 66.71 & \textbf{55.56} & \textbf{31.76} & \textbf{55.31} \\
 & 2.40 & \textbf{2.81} & \textbf{6.52} & \textbf{6.31} & 6.92 & \textbf{9.36} & \textbf{5.82} & \textbf{6.33} & \textbf{5.42} & \textbf{5.69} \\ \hline

\end{tabular}
\end{table*}
\subsubsection{Highway Scenarios}
The localization results for highway scenarios are summarized in Table~\ref{tab:Highway}. It is evident that VIO systems face significant challenges in such environments, primarily due to two factors. First, reliable initialization under high-speed motion is inherently difficult for VIO. Although methods such as VINS-Mono and OpenVINS provide dynamic initialization, they still require slow motion to achieve accurate initialization. In contrast, VIWO systems utilize wheel measurements to enable robust initialization. Secondly, highway motion is typically dominated by uniform linear movement, which induces degenerate conditions for VIO~\cite{c4}. This leads to additional unobservable degrees of freedom and ultimately degrades the estimation accuracy of VIO systems.

Among the VIWO systems, PL-VIWO2 consistently outperforms VIW-Fusion and MINS across most test sequences in both monocular and stereo configurations. It ranks first in 8 of 11 sequences and second in the remaining 3 under the monocular setup, and first in 10 of 11 sequences and second in 1 under the stereo setup. This superior performance can be attributed to its ability to account for dynamic features and the incorporation of additional visual constraints from line features. Notably, VIW-Fusion exhibits large angular errors on some sequences, such as Urban20 and Urban21. This is mainly because VIW-Fusion relies on a global plane motion assumption enforcing a zero z-axis offset, which does not hold for all datasets and leads to significant rotation errors. In contrast, PL-VIWO2 applies a plane constraint based on a local reference plane, making it more robust to variations in vertical motion. Fig.~\ref{fig:urban36} shows an example of trajectories generated by different algorithms on Urban36. As illustrated, stereo PL-VIWO2 achieves the closest alignment with the ground truth among all methods.

\subsubsection{City Scenarios}
Table.~\ref{tab:Urban} summarizes the localization results in urban scenarios. Compared with highway navigation, urban driving is generally slower and provides richer texture information, enabling VIO systems to achieve better performance. Overall, stereo-based VIO and VIWO algorithms tend to deliver higher accuracy than their monocular counterparts. For monocular VIO methods, performance often degrades under the constant velocity motion pattern, a typical degenerate case~\cite{c5}, and in dynamic environments such as traffic scenes with other moving vehicles. In contrast, monocular VIWO systems benefit from integrating wheel measurements, which provide additional motion constraints and enhance robustness in such challenging conditions. Notably, in the Urban34 sequence, most VIO methods fail due to excessive lighting variations, whereas PL-VIO remains robust owing to the incorporation of additional line feature constraints.

Compared with MINS, PL-VIWO2 achieves further improvements in both monocular and stereo configurations by incorporating a larger number of feature points along with additional line features. Monocular VIW-Fusion achieves the best results in several sequences thanks to its optimization-based framework, which improves robustness at the cost of higher computational complexity. Overall, monocular PL-VIWO2 ranks first in 5 out of 10 sequences and second in the remaining 5, while stereo PL-VIWO2 ranks first in 7 out of 10 sequences and second in 2. As illustrated in Fig.~\ref{fig:urban30}, which plots the trajectory estimates of different algorithms alongside the ground truth, stereo PL-VIWO2 produces results most closely aligned with the reference trajectory.

\subsection{Ablation Study}
To demonstrate the effectiveness of the proposed line process methods, we conduct an ablation study by first comparing the results with the MINS and our previous PL-VIWO~\cite{c16}, and then removing key modules from PL-VIWO2 — including line feature update, and line optimization refinement. Since PL-VIWO only support monocular setup, we evaluate the localization accuracy in this configuration after removing each component, in order to quantify individual contributions. 

The ablation study results on the KAIST Urban dataset are summarized in Table.~\ref{tab:ablation} and the best results are \textbf{highlighted}. Compared with PL-VIWO, PL-VIWO2 achieves further improvements in localization accuracy owing to the newly introduced line tracking and optimization refinement modules. Compared with MINS, PL-VIWO2 without line features additionally incorporates the MCC module, which ensures that more valid feature points can be used for state updates, thereby improving accuracy. Furthermore, when line features are included, PL-VIWO2 shows significant improvements across most sequences, further demonstrating the effectiveness of incorporating line features.

\subsection{Computation Efficiency Analysis}
To demonstrate the runtime efficiency of the proposed system, we first record the execution time of each individual feature-processing module in PL-VIWO2 under two representative scenarios and compare it with PL-VINS. We then evaluate the overall state update time and CPU utilization against those of state-of-the-art methods.

\subsubsection{Feature Processing Time}
\begin{table}
    \centering
    \caption{Average Processing Time of Principal Component with PL-VIWO2 in Urban38.}
    \resizebox{0.49\textwidth}{!}{
    \begin{tabular}{lcccccc}
        \toprule
        & \makecell[c]{Point \\ Extraction} 
        & \makecell[c]{Line \\ Extraction} 
        & \makecell[c]{Line \\ Classification}
        & \makecell[c]{Point \\ Matching}
        & \makecell[c]{Line \\ Matching} 
        & \makecell[c]{Total \\ Time}  \\
        \midrule
        PL-VIWO2 & 5 ms & 7 ms & 0.1 ms & 2.5 ms & 0.9 ms & 15.5 ms  \\
        PL-VINS  & 10 ms & 23 ms & 0 ms & 3.0 ms & 6 ms & 42 ms \\
        \bottomrule
    \end{tabular}}
    \label{tab:feature-time}
\end{table}

\begin{figure}[t]
    \centering
    \subfloat{
        \includegraphics[width=1.0\linewidth]{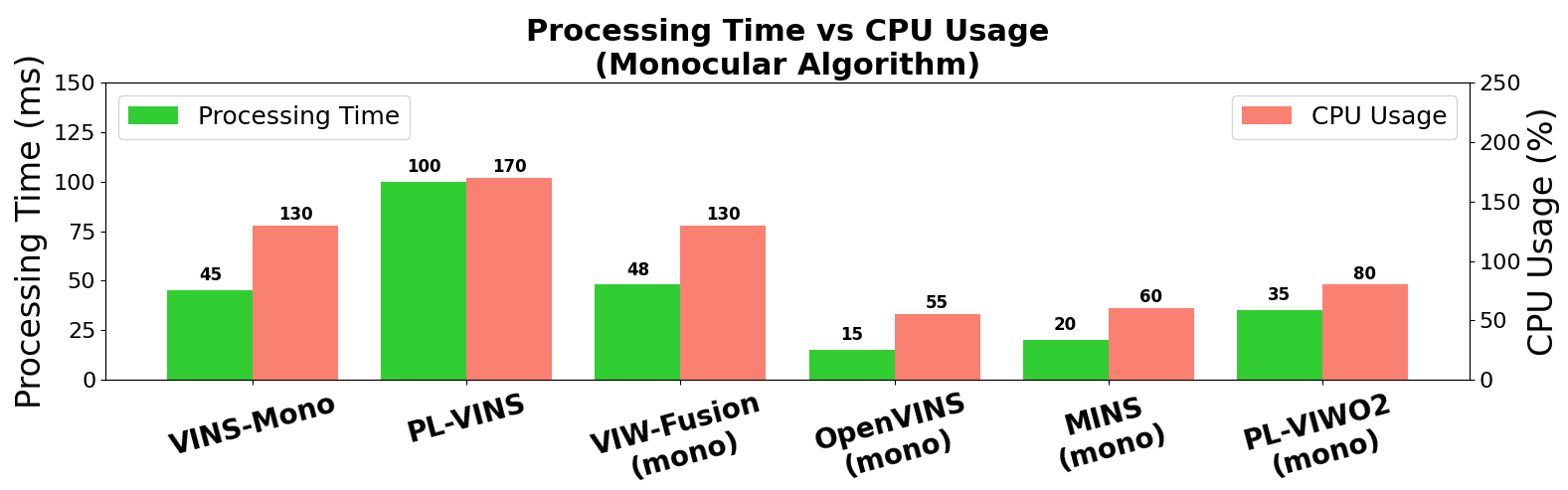}} \\
    \subfloat{
        \includegraphics[width=1.0\linewidth]{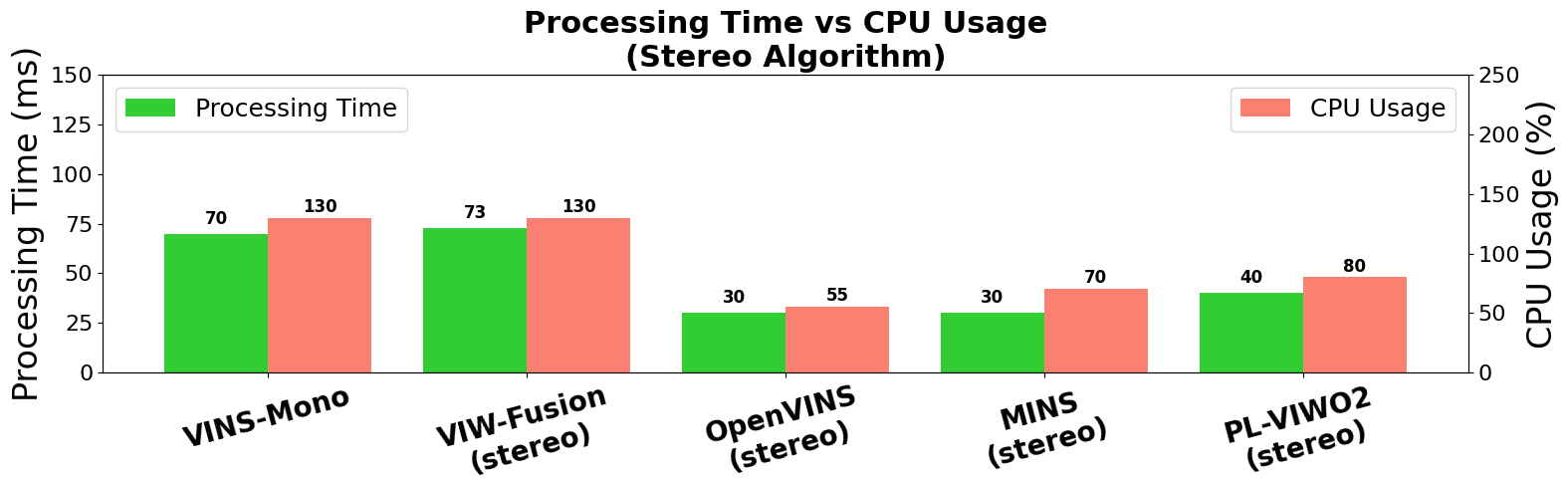}}
    \caption{Bar charts of average processing time (s) between frames and CPU usage (\%) for monocular algorithms and stereo algorithms on Urban28.}
    \label{fig:process-time}
\end{figure}

\begin{figure}[!ht]
    \centering
    \subfloat{\includegraphics[width=0.15\textwidth, trim=30 30 30 30,clip]{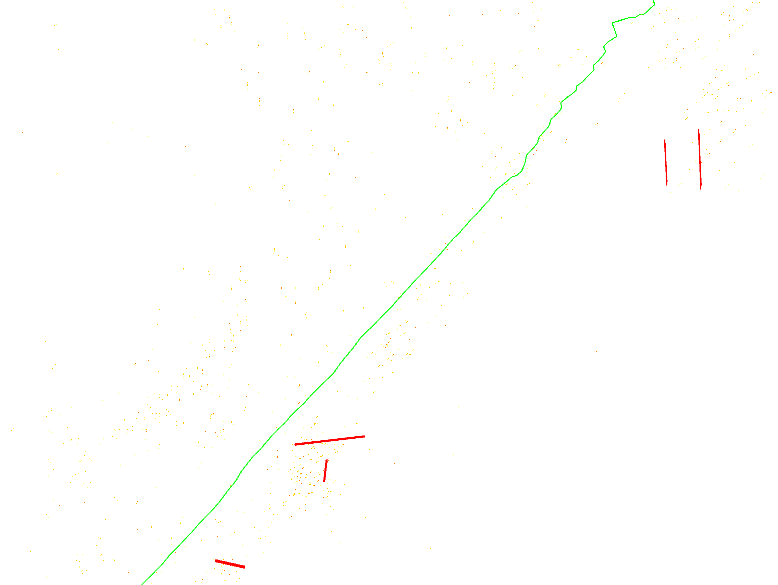}}
    \subfloat{\includegraphics[width=0.15\textwidth, trim=30 30 30 30,clip]{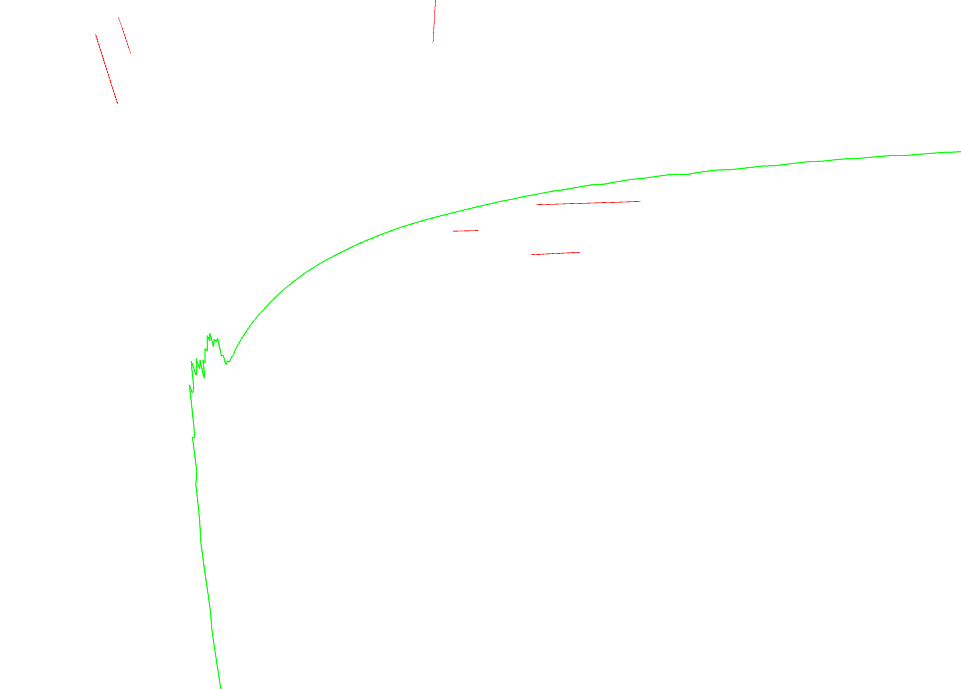}}
    \subfloat{\includegraphics[width=0.15\textwidth, trim=30 30 30 30,clip]{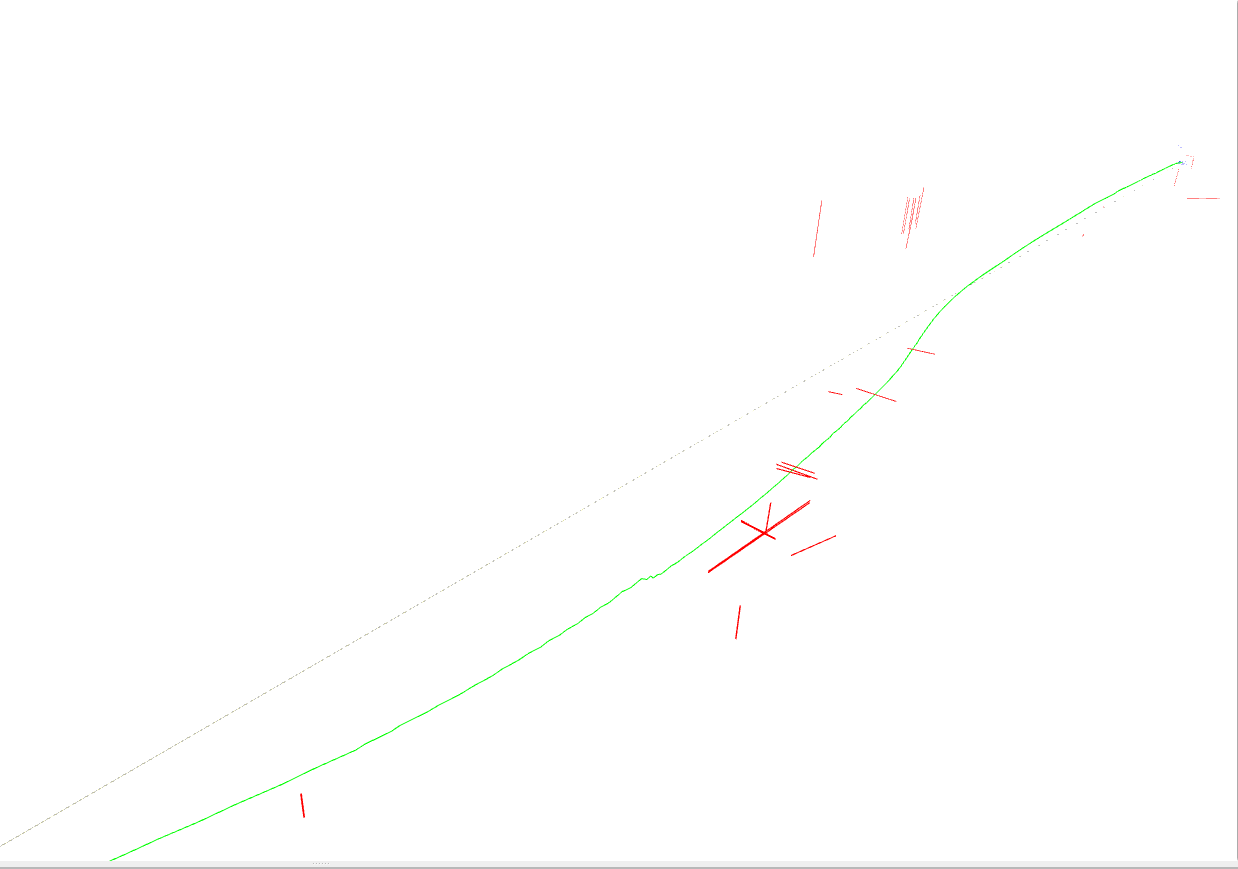}}
    \\
    \subfloat{\includegraphics[width=0.15\textwidth, trim=30 30 30 30,clip]{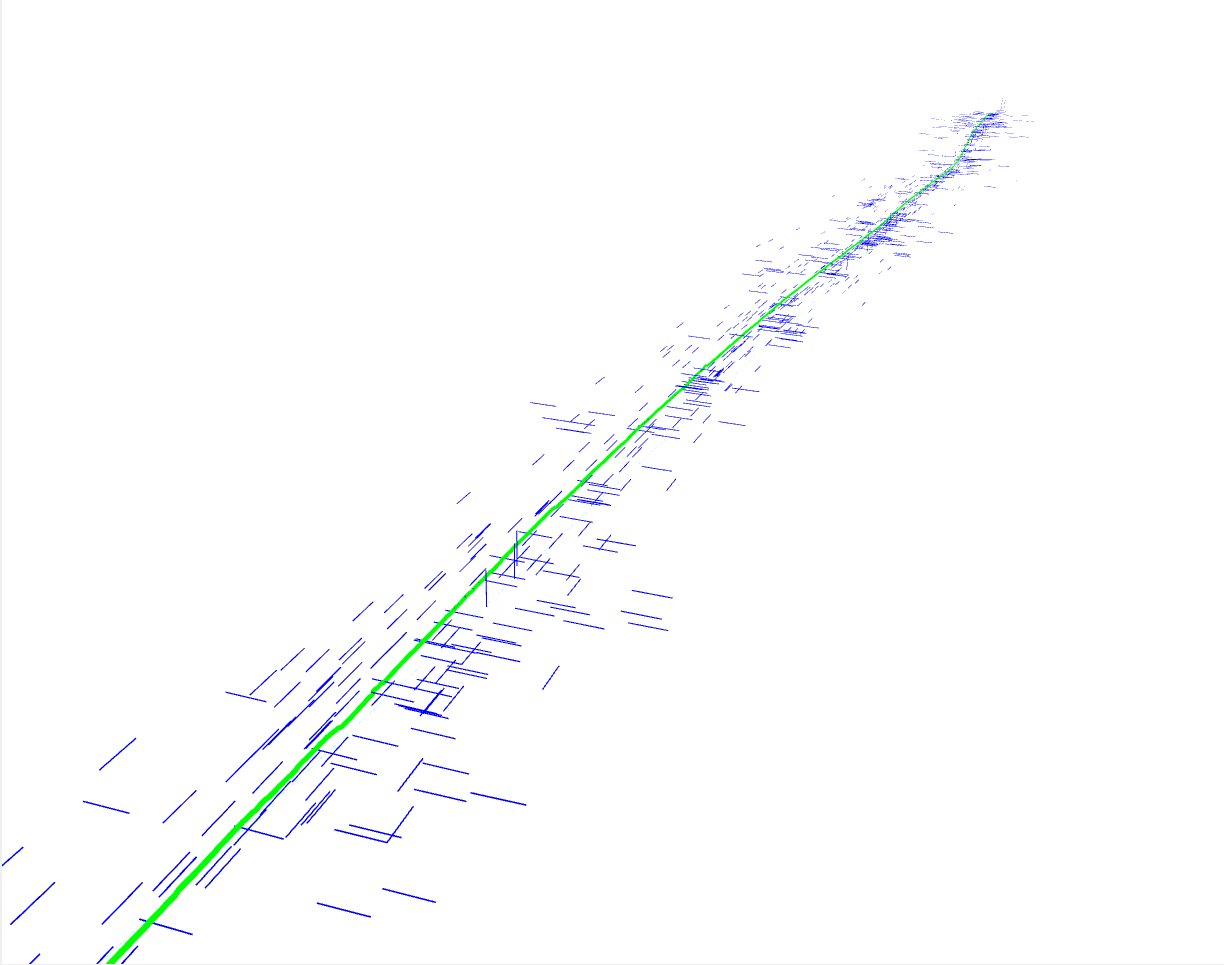}}
    \subfloat{\includegraphics[width=0.15\textwidth, trim=30 30 30 30,clip]{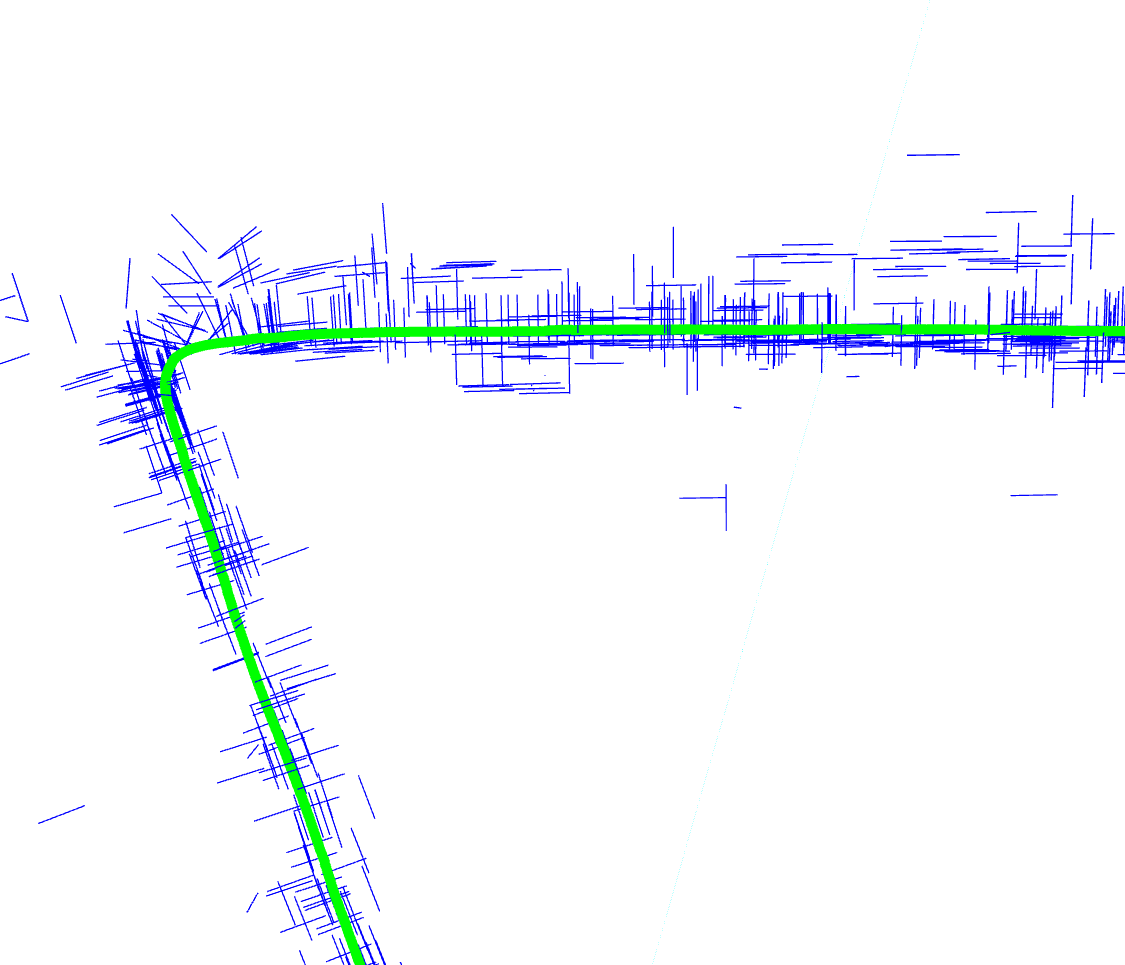}}
    \subfloat{\includegraphics[width=0.15\textwidth, trim=30 30 30 30,clip]{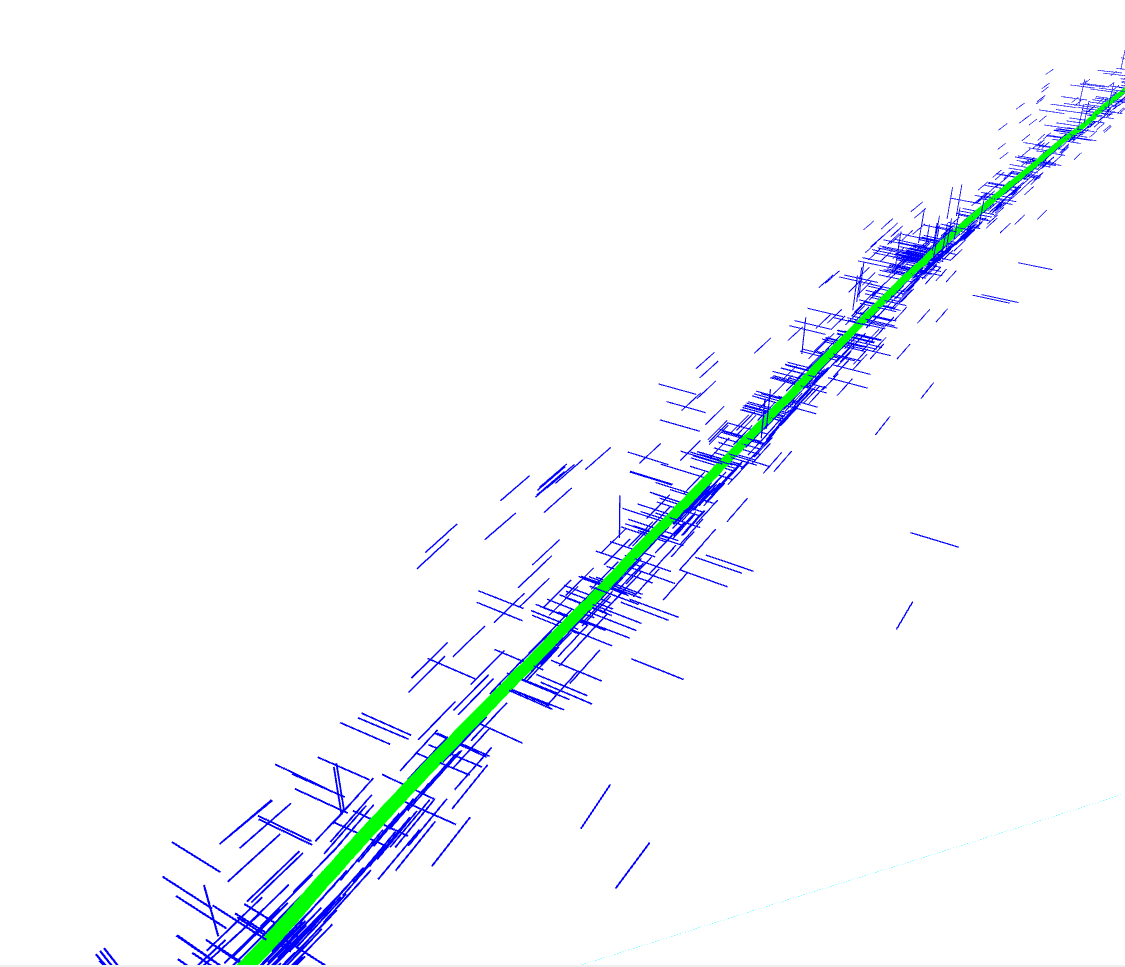}}
    \caption{Comparison of line triangulation results across various sequences. The top row shows the results of PL-VINS (red lines), while the bottom row presents those of PL-VIWO2 in the monocular setup (blue lines).
    }
    \label{fig:Triangulatuon}
\end{figure}
Table.~\ref{tab:feature-time} summarizes the average runtime of each image processing module between adjacent frames in the monocular setup of PL-VIWO2, providing a direct comparison with PL-VINS. As shown in the table, our system achieves noticeably lower processing times for both point and line feature processing. For point feature extraction and matching, the improvement mainly stems from the grid-based point extraction inherited from OpenVINS, which ensures a uniform distribution of feature points across the image and enables faster, more efficient tracking. For line feature extraction, the FLD delivers significantly higher efficiency compared to the modified LSD algorithm implemented in PL-VINS. As for line matching, although we adopt a two-stage tracking strategy, the overall computational cost remains low thanks to the initial point-line association, which effectively reduces redundant matching operations. Additionally, the line classification step incurs minimal computational overhead. Consequently, the total feature processing time of our system is much lower than PL-VINS.
\begin{figure*}[!ht]
    \centering
    \subfloat[urban26]{\includegraphics[width=0.15\textwidth, trim=10 10 10 10,clip]{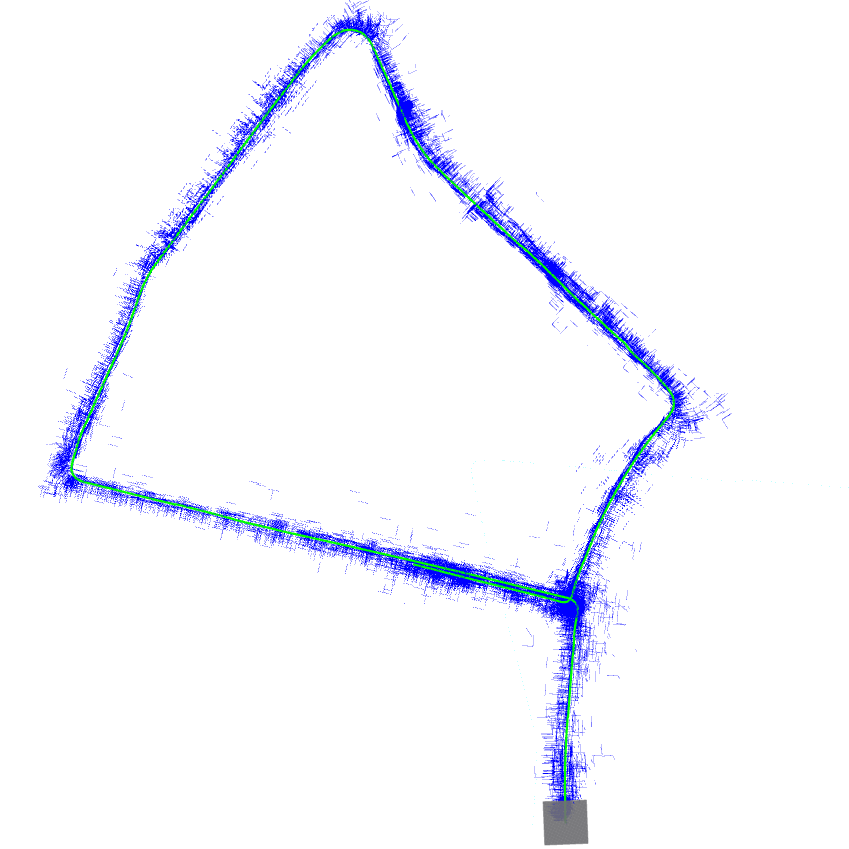}}
    \hspace{0.03\textwidth}
    \subfloat[urban27]{\includegraphics[width=0.15\textwidth, trim=10 10 10 10,clip]{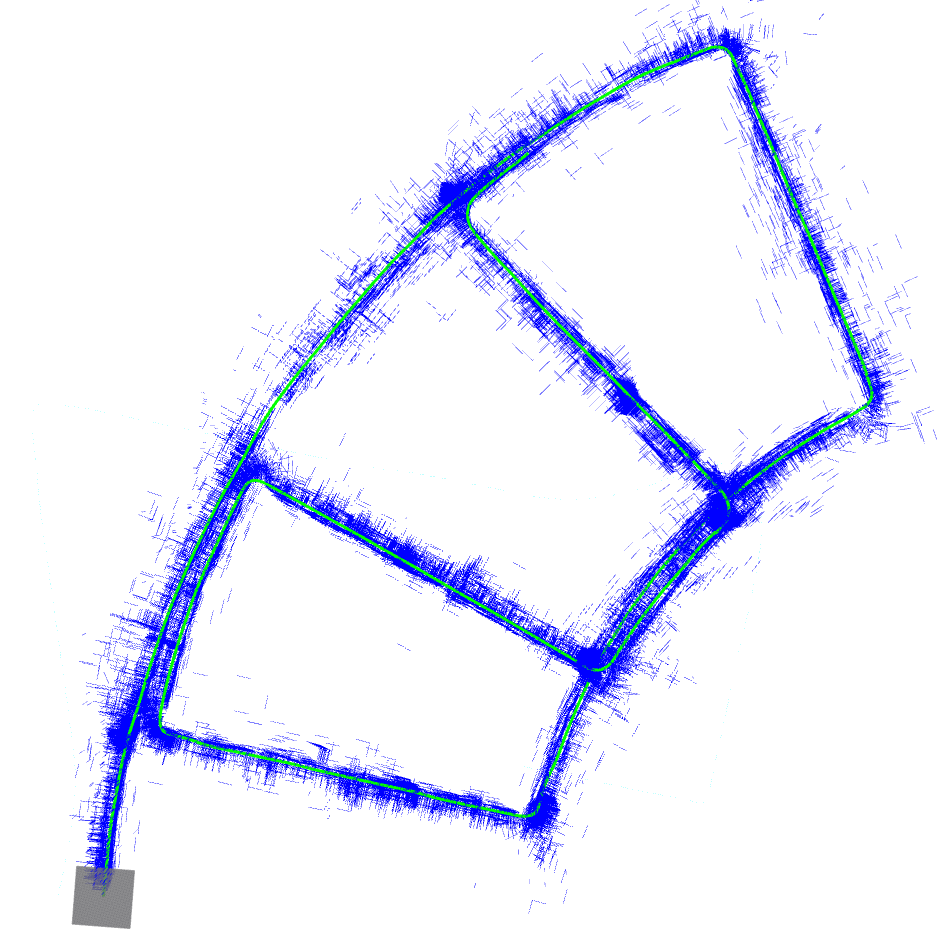}}
    \hspace{0.03\textwidth}
    \subfloat[urban28]{\includegraphics[width=0.15\textwidth, trim=10 10 10 10,clip]{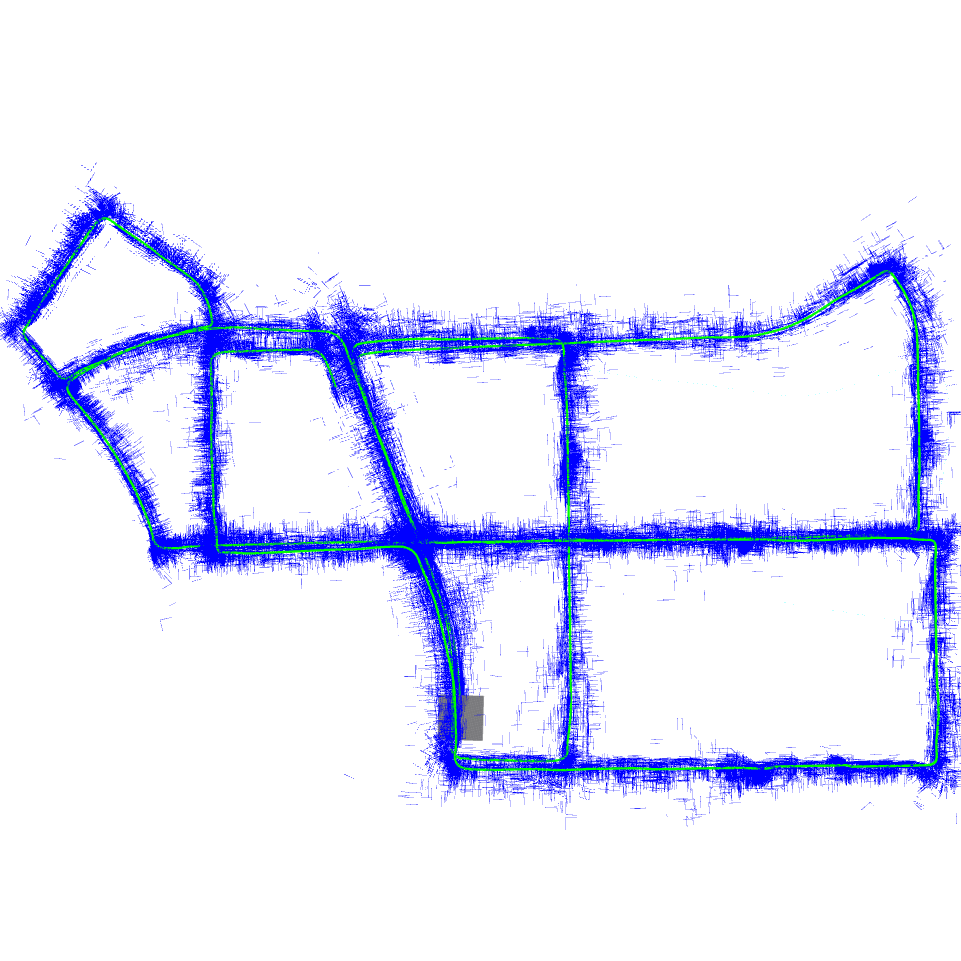}}
    \hspace{0.03\textwidth}
    \subfloat[urban29]{\includegraphics[width=0.15\textwidth, trim=10 10 10 10,clip]{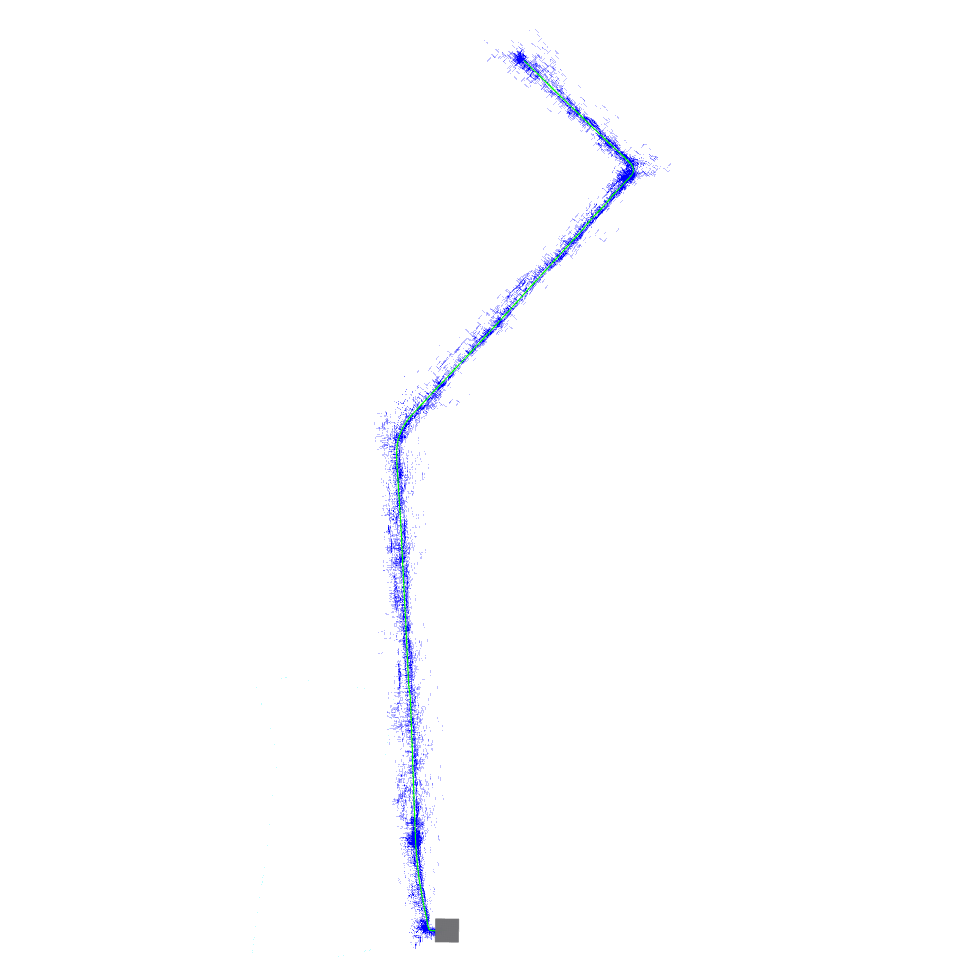}}
    \hspace{0.03\textwidth}
    \subfloat[urban30]{\includegraphics[width=0.15\textwidth, trim=10 10 10 10,clip]{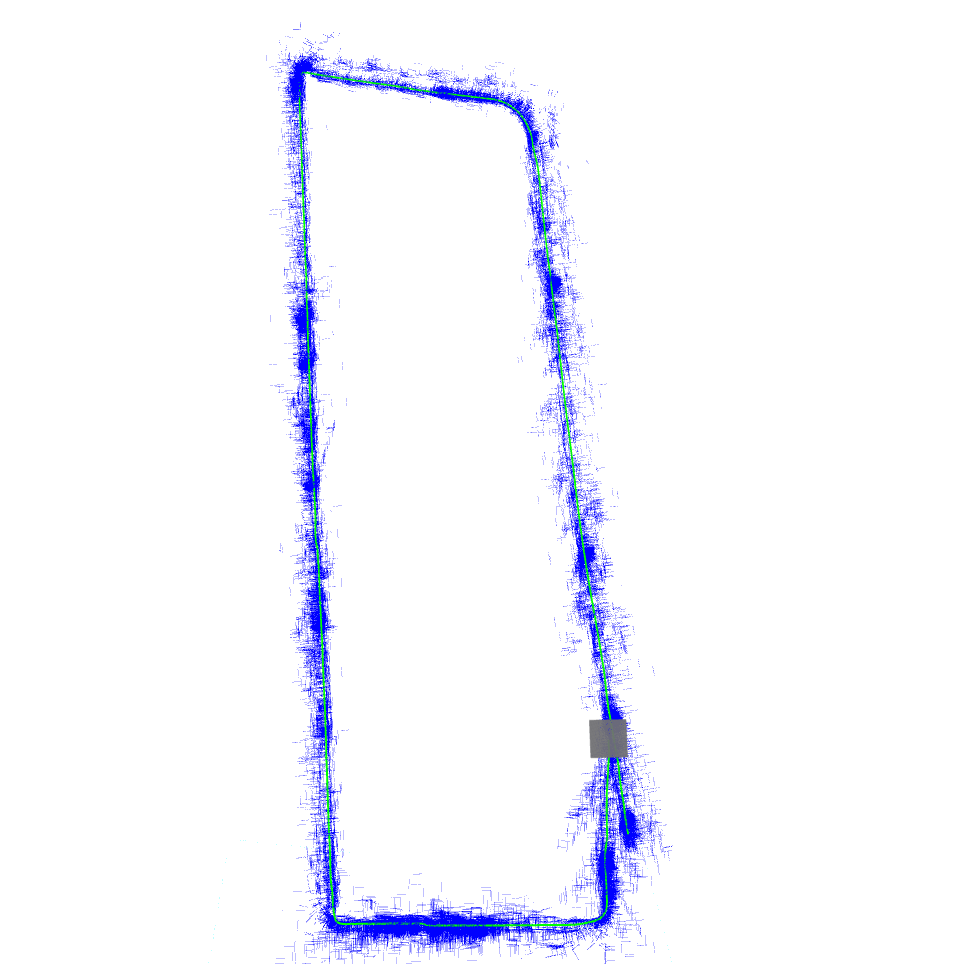}}
    \hspace{0.03\textwidth}
    \\
    \subfloat[urban31]{\includegraphics[width=0.15\textwidth, trim=10 10 10 10,clip]{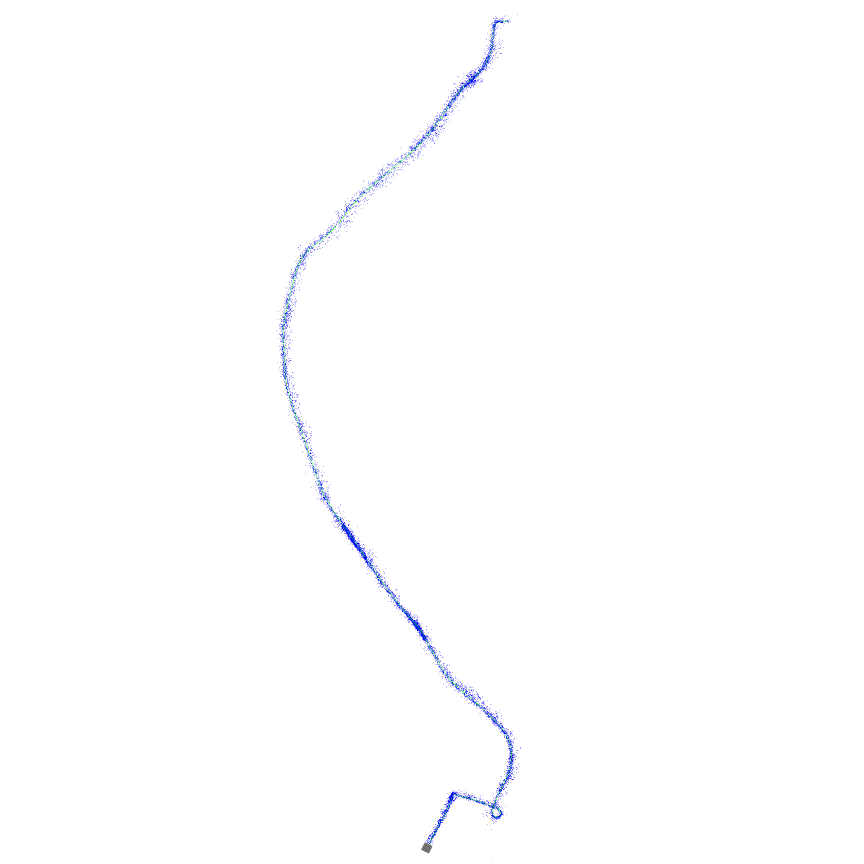}}
    \hspace{0.03\textwidth}
    \subfloat[urban32]{\includegraphics[width=0.15\textwidth, trim=10 10 10 10,clip]{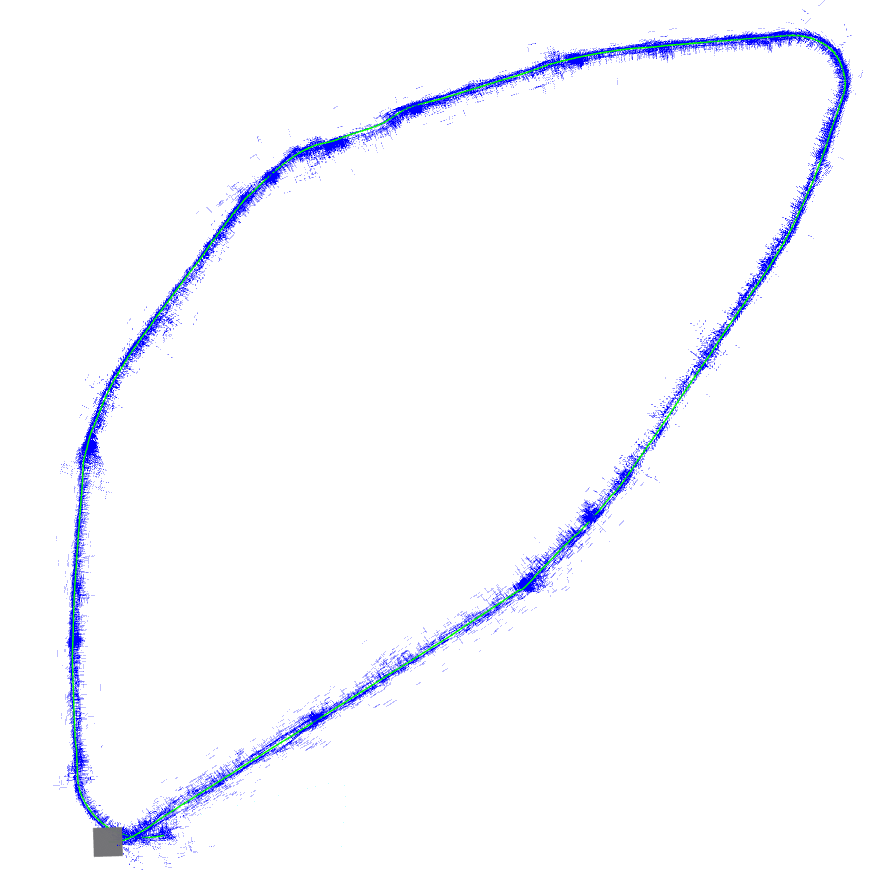}}
    \hspace{0.03\textwidth}
    \subfloat[urban33]{\includegraphics[width=0.15\textwidth, trim=10 10 10 10,clip]{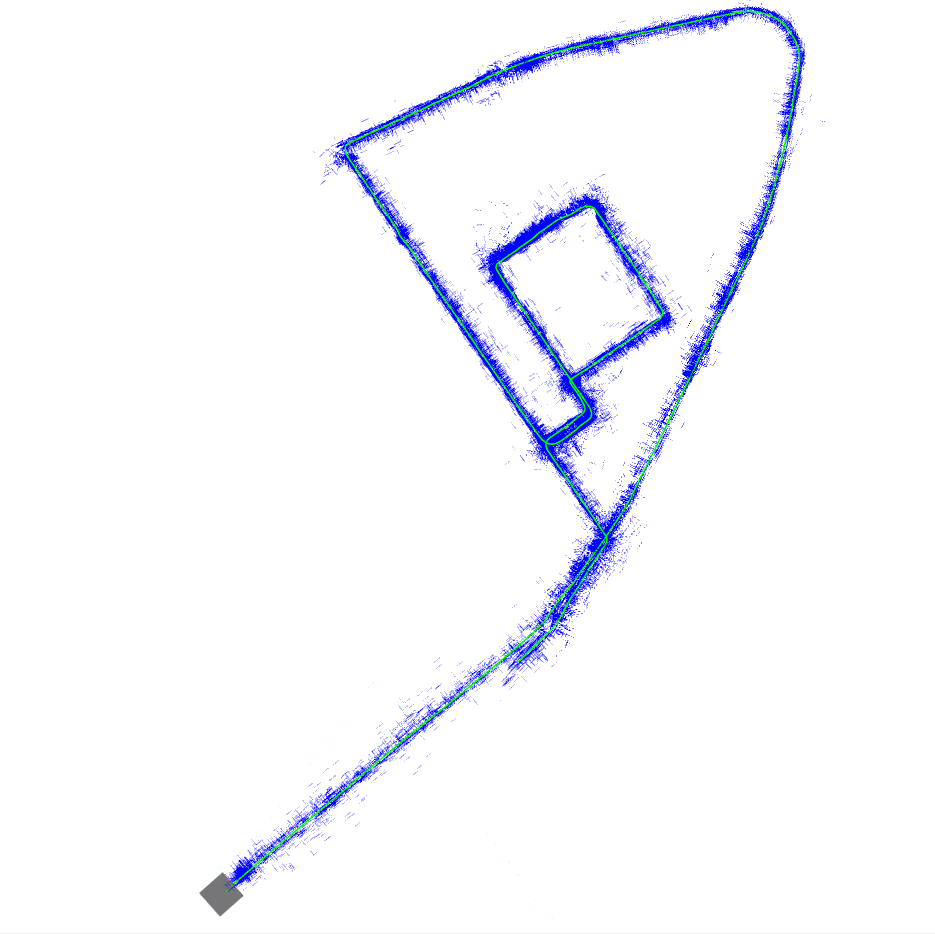}}
    \hspace{0.03\textwidth}
    \subfloat[urban34]{\includegraphics[width=0.15\textwidth, trim=10 10 10 10,clip]{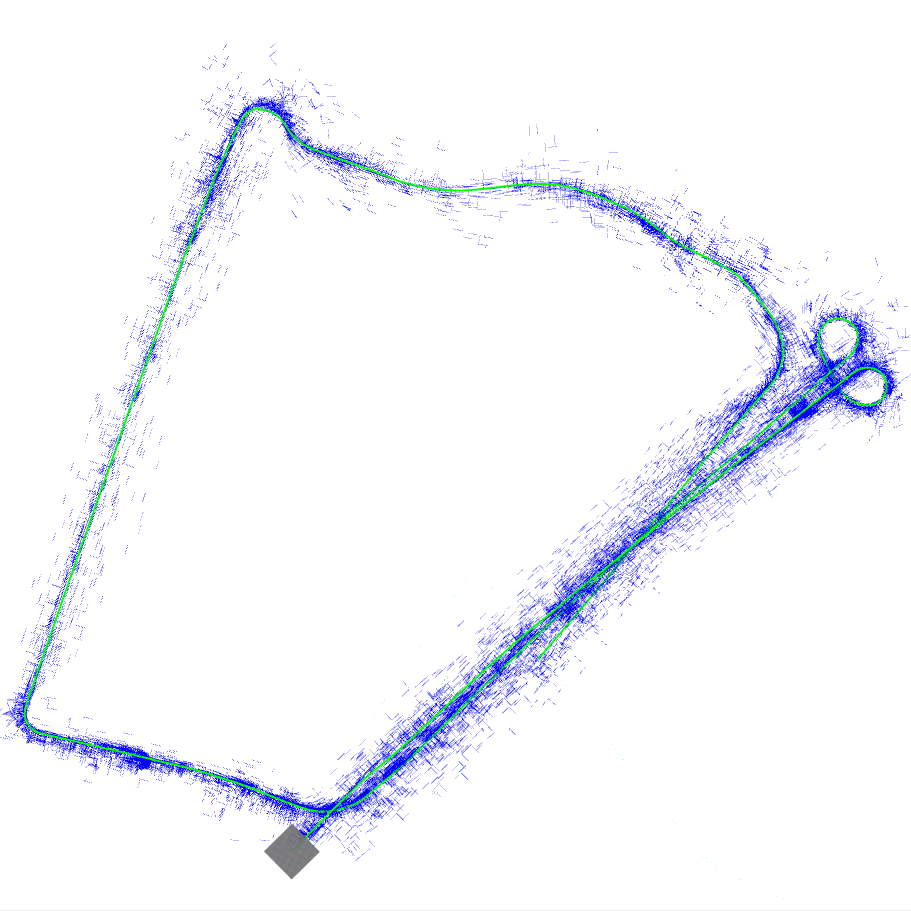}}
    \hspace{0.03\textwidth}
    \subfloat[urban38]{\includegraphics[width=0.15\textwidth, trim=10 10 10 10,clip]{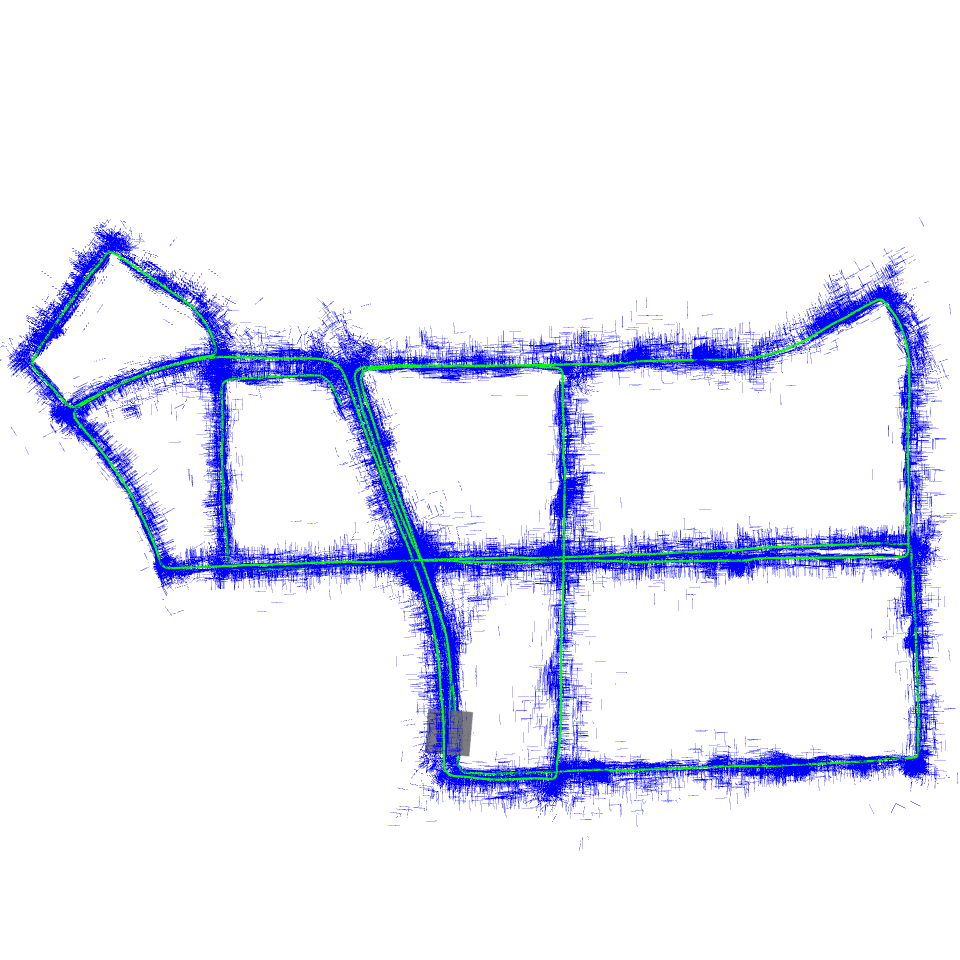}}
    \hspace{0.03\textwidth}
    \caption{Estimated trajectories (green) with lines (blue) obtained by PL-VIWO2 under the stereo configuration in city driving sequences.}
    \label{fig:trajs}
\end{figure*}
\subsubsection{Update Time$\And$CPU Usage} 
We then evaluate the average processing time per update and the CPU usage of each algorithm under both monocular and stereo configurations, as shown in Fig.~\ref{fig:process-time}. For filter-based algorithms, the processing time is measured for each visual update step, whereas for optimization-based algorithms, the average image processing time between two keyframes together with the optimization time is recorded.

Overall, optimization-based algorithms require additional computation and threads to iteratively solve the factor graph optimization problem, leading to a significant increase in both processing time and CPU usage compared with filter-based algorithms. OpenVINS achieves the lowest processing time and CPU usage in both configurations, as it employs only two sensors and processes solely point features. MINS slightly increases processing time by incorporating wheel measurements, while CPU usage remains unchanged. In contrast, PL-VIWO2 maintains low computational overhead while incorporating line features to improve accuracy, making it particularly suitable for long-term accurate and robust localization on resource-constrained edge devices. 

\subsection{Line Triangulation Analysis}
To demonstrate the effectiveness of the proposed line triangulation methods, we compare the monocular configuration of PL-VIWO2 against PL-VINS across different sequences, as illustrated in Fig.~\ref{fig:Triangulatuon}. It is evident that only a limited number of lines are successfully triangulated by PL-VINS, which can be attributed to two main factors. First, the descriptor-based line tracking method exhibits low efficiency in complex outdoor complex environments. Second, many lines fail to be reliably triangulated due to the degenerate motion that frequently occurs during driving. In contrast, PL-VIWO2 overcomes these limitations, enabling a significantly larger number of lines to be successfully triangulated and subsequently utilized for state updates. To further demonstrate the robustness and accuracy of line triangulation in PL-VIWO2, multiple trajectories incorporating 3D line features are plotted for urban sequences, as shown in Fig.~\ref{fig:trajs}. 

\section{Conclusion}
\label{sec:conclusion}
To enable stable and accurate localization for long-term outdoor urban navigation, we propose PL-VIWO2, a VIWO system that jointly leverages points and lines. It features a novel point–line processing pipeline for efficient tracking and triangulation, an $\mathrm{SE}(2)$-constrained $\mathrm{SE}(3)$ wheel pre-integration to enforce planar motion constraints, and a motion consistency check (MCC) to filter dynamic features. Extensive simulations and real-world experiments confirm its accuracy and robustness. Nonetheless, the system still depends on reliable point or line extraction and matching, which may degrade under low illumination or high-speed motion. Future work will explore learning-based feature extraction and matching to further improve robustness.

\newpage

 




\vfill

\end{document}